%% file: ICML2026.tex
\documentclass{article}

\usepackage{microtype}
\usepackage{graphicx}
\usepackage{subcaption}
\usepackage{booktabs} 
\usepackage{multirow}
\usepackage{pifont}
\usepackage{hyperref}

\usepackage[preprint]{icml2026}

\usepackage{amsmath}
\usepackage{amssymb}
\usepackage{mathtools}
\usepackage{amsthm}

\usepackage[capitalize,noabbrev]{cleveref}

\theoremstyle{plain}

\theoremstyle{definition}

\theoremstyle{remark}

\usepackage[textsize=tiny]{todonotes}

\usepackage[most]{tcolorbox}
\usepackage{amssymb}
\renewcommand{\paragraph}[1]{\vspace{1.25mm}\noindent\textbf{#1}}

\icmltitlerunning{Submission and Formatting Instructions for ICML 2026}

\begin{document}

\twocolumn[
  \icmltitle{Evaluating and Steering Modality Preferences in Multi-modal LLMs}



  \icmlsetsymbol{equal}{*}

  \begin{icmlauthorlist}
    \icmlauthor{Yu Zhang}{1,2}
    \icmlauthor{Jinlong Ma}{1}
    \icmlauthor{Yongshuai Hou}{2}
    \icmlauthor{Xuefeng Bai}{1}
    \icmlauthor{Kehai Chen}{1,2}
    \icmlauthor{Yang Xiang}{2}
    \icmlauthor{Jun Yu}{1}
    \icmlauthor{Min Zhang}{1,2}

  \end{icmlauthorlist}

  \icmlaffiliation{1}{Harbin Institute of Technology, Shenzhen, China}
  \icmlaffiliation{2}{Peng Cheng Laboratory, Shenzhen, China}

  \icmlcorrespondingauthor{Yu Zhang}{yuzhang2717@gmail.com}
  \icmlcorrespondingauthor{Kehai Chen}{chenkehai@hit.edu.cn}

  \icmlkeywords{Machine Learning, ICML}

  \vskip 0.3in
]



\printAffiliationsAndNotice{}  

\input{sec/abs}
\input{sec/intro}
\input{sec/related}

\input{sec/dataset}

\input{sec/exp}

\input{sec/method}
\input{sec/conclusion}

\section*{Impact Statement}
The proposed MC\textsuperscript{2} evaluates the modality preference of MLLMs. 
Understanding which modality a model prioritizes could be used to circumvent safety mechanisms (e.g., hiding harmful content in the favored modality), making it harder for filters to detect inappropriate content.
Therefore, it is essential to incorporate effective safeguards in MLLMs to filter out any inappropriate materials.
\bibliography{example_paper}

\input{sec/supp}



\bibliographystyle{icml2026}



\end{document}

%% file: sec/abs.tex
\begin{abstract}

Multi-modal large language models (MLLMs) have achieved remarkable success on complex multi-modal tasks. 
However, it remains insufficiently explored whether they exhibit \textit{modality preference}, a tendency to favor one modality over another when processing multi-modal contexts.
To study this question, we introduce \textbf{MC\textsuperscript{2}} benchmark, which constructs controlled evidence-conflict scenarios to systematically evaluate modality preference in decision-making.
Extensive experiments reveal that all 20 tested MLLMs generally demonstrate clear modality preferences, and such preferences can serve as a useful indicator of downstream task performance of MLLMs. 
Further analysis shows that modality preference can be controlled by instruction guidance and captured within the latent representations of MLLMs.
Built on these insights, we propose a probing and steering method based on representation engineering to explicitly control modality preference without requiring additional fine-tuning. 
This method effectively amplifies modality preference toward a desired direction and demonstrates promising improvements across 
multiple multi-modal understanding and reasoning tasks.
\begingroup
\renewcommand{\thefootnote}{}
\footnotetext{The code and data are released at \url{https://github.com/EchoDreamer/Modality-Preference}.}
\addtocounter{footnote}{-1}
\endgroup
\end{abstract}

%% file: sec/intro.tex
\section{Introduction}
\label{sec:intro}
multi-modal Large Language Models (MLLMs;~\citealp{gpt4,gemini,qwen2vl,survey1}) have emerged as a powerful paradigm for processing and reasoning across heterogeneous data modalities (e.g., text, images, video). 
Recent advances demonstrate their exceptional capabilities on complex tasks with multi-modal contexts, including autonomous web browsing~\citep{2}, graphical user interface understanding~\citep{1}, and multi-modal dialogue systems~\citep{dialogue}. 
Despite impressive performance, fundamental questions remain about their \textit{modality preference}—whether MLLMs tend to rely more heavily on one modality than others, and to what extent they favor a specific modality when resolving multi-modal inputs.

\begin{figure}[t]
 \centering
 \small
  \includegraphics[width=0.4\textwidth]{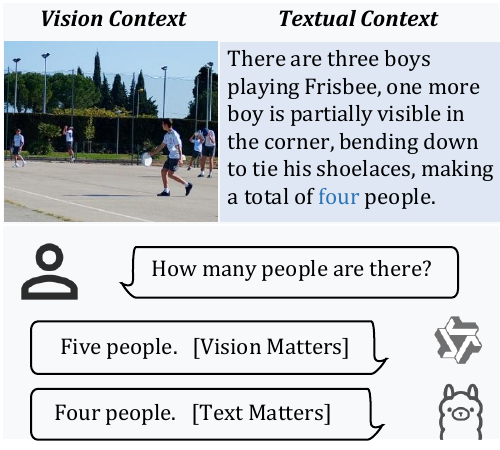}
  \caption{Illustrations of evaluating modality preference. 
  Using multi-modal conflict context pairs to evaluate modality preference.}
  \label{fig:motivation}
\end{figure}
To investigate this, one line of work~\citep{fu2024isobench,amara2024context} compares model performance on unimodal input, providing either only text or only image input for the same question.
Another line of research analyzes the relative contributions of textual and visual context, typically by removing one modality to observe the changes in the downstream performance~\citep{park2025assessing} or Shapley value~\citep{alishahi2019analyzing,parcalabescu2024vision,parcalabescu2022mm}.
However, such settings inherently introduce bias, as \textit{they isolate modalities, thus failing to reflect how models process inputs in realistic multi-modal scenarios}, where information from different modalities naturally co-occurs.

In this paper, we provide a controllable setup to study the modality preference in MLLMs. 
As shown in the Figure~\ref{fig:motivation}, we introduce a modality context conflict setting, where  MLLMs are asked to answer a question based on a pair of contrasting pieces of evidence from different modalities.
In this way, we can determine modality preference based on the answer given by MLLMs.

To enable a rigorous and fair assessment, we eliminate existing confounding factors including the internal knowledge of MLLMs and inherent understanding capabilities. 
Consequently, we curate and annotate the dataset to ensure that the model achieves precise question and single-modality context comprehension.  
Building upon this, we introduce a semi-automated annotation framework to construct a refined \textbf{M}odality \textbf{C}ontext \textbf{C}onflict dataset, \textbf{\textsc{MC\textsuperscript{2}}}, which covers eight tasks with 2,000 carefully selected samples. 
Using MC\textsuperscript{2}, we conduct a comprehensive analysis of modality preference across a diverse set of 20 representative MLLMs. 
Our study reveals several intriguing findings:
\begin{itemize}
    \item Most MLLMs (except Qwen2.5-VL and InternVL3) display text preference, as shown in the Figure~\ref{fig:single}, and modality preference can serve as a useful indicator of downstream task performance.
    
    \item Internal attention patterns toward specific modalities give rise to modality preference, and the underlying factors can be traced to the training data recipe and the scale of the MLLMs.
    
    \item Modality preference can be modulated through explicit instruction guidance, and the direction of preference can be captured as geometrically separable patterns in the latent space.
\end{itemize}

Built on these, we propose a modality preference probing and steering method based on representation engineering~\citep{zou2023representation} to explicitly amplify the modality preference without additional fine-tuning. 
Experimental results show that the proposed method leads to notable performance improvements on multi-modal understanding and reasoning tasks.
Our main contributions are summarized as follows:

\begin{itemize}
    \item We introduce \textbf{\textsc{MC\textsuperscript{2}}} to comprehensively evaluate modality preferences in MLLMs and highlight the significance of modality preference in correlating the downstream task performance.
    \item Our analysis reveals that intrinsic modality preferences in MLLMs are steerable and identifiable through latent representations, providing insights into multi-modal reasoning.
    \item We propose a training-free method that steers modality preference via representation-level intervention, enabling controllable preference adjustment and enhancing performance on downstream tasks.
\end{itemize}

%% file: sec/related.tex
\section{Related Work}
\subsection{multi-modal Large Language Models}
In recent years, Large Language Models (LLMs) have achieved impressive results across a wide spectrum of tasks~\citep{Advances,zhang-etal-2024-question,guo2024baner,xu2025memory,Reasoning}. 
Building upon these, multi-modal Large Language Models (MLLMs) have extended such capabilities, demonstrating strong performance in visual question answering, structured information understanding, and visual navigation~\citep{zhu2024benchmarking,zhang2025momakitchen,liu2025bridge,wei2023instructiongpt,zhang2025will}.
Recent research has turned to systematically evaluating their biases and failure modes~\citep{phd,wei2024diff,amara2024context,alishahi2019analyzing,chen2024quantifying}, aiming to better understand their limitations and guide the development of more reliable models.

\subsection{Modality Preference}
Existing studies on the modality preference of MLLMs can be broadly divided into two categories: 1) investigating the data-related factors that give rise to modality preference or bias, and 2) analyzing the intrinsic characteristics of modality preference within models.

\noindent\textbf{Data factors influencing modality preference.}
Research on data-related factors~\citep{b_bias,b_bias1_1,b_bias2_1} explores how properties of multi-modal datasets give rise to and reinforce modality preference.
In particular, many samples in multi-modal datasets can be resolved correctly by relying on information from only a single modality. 
When prevalent in training data, such samples bias the optimization dynamics, encouraging models to disproportionately rely on a single modality~\citep{b_bias1_1}.
Furthermore, their inclusion in evaluation benchmarks artificially inflates performance metrics, as models can exploit these unimodal shortcuts instead of performing genuine cross-modal integration~\citep{b_bias2_1,b_bias2_2}.
While these studies establish data's role in inducing preference, our work focuses on exploring the intrinsic modality preference inherent in MLLMs themselves, independent of specific data distributions.

\noindent\textbf{Evaluating the intrinsic modality preference in MLLMs.}
Early studies~\citep{bias1,bias2,bias3} analyze modality preference or bias by examining how multi-modal models optimize for multi-modal inputs.
Through such analyses, researchers observe that modality bias has a significant impact on both model optimization and downstream task performance~\citep{bias1,bias4,bias5}.
While these works offer valuable insights, they typically require training models from scratch, which makes them impractical for large-scale multi-modal systems.
Recent studies have investigated intrinsic modality preference in MLLMs by evaluating model performance on unimodal inputs—using only text or only image for the same task~\citep{fu2024isobench,amara2024context}—and by applying Shapley value-based attribution methods to quantify the contribution of each modality~\citep{alishahi2019analyzing,parcalabescu2022mm,parcalabescu2024vision}.
However, in real-world multi-modal applications, all modalities are indispensable for task resolution, making these frameworks \textit{inadequate for determining true modality preference}. 
\cite{wu2025foundation} evaluate the model’s ability to detect conflict under scenarios involving conflicting multi-modal contexts. 
However, conflict detection is only one facet of multi-modal reasoning and does not comprehensively reflect a model’s modality preference when processing multi-modal contexts.

In this work, we investigate the behavioral tendencies of MLLMs by examining their responses to questions within scenarios involving conflicting multi-modal contexts.
Unlike prior studies, we decouple modality preference from confounding variables—such as internal model knowledge and varying understanding capabilities—by ensuring the models can accurately comprehend the provided context.
Furthermore, we design a flexible method to steer the modality preference and demonstrate effectiveness across multiple downstream tasks.

\subsection{Representation Engineering} 

Extensive research has shown that large language models (LLMs) encode interpretable concepts, such as sentiment, truthfulness, and stylistic attributes in representation space in LLMs~\citep{steering4,steering1,steering2,steering3}. 
Building on this foundation, representation engineering has proven effective for editing, enhancing, or suppressing specific behaviors in LLMs~\citep{represent1,stolfo2024improving,represent3,represent2,zou2023representation}. 
In this work, we extend this paradigm to a novel setting: controlling modality preference in multi-modal large language models (MLLMs). 
Instead of focusing on abstract properties, our method identifies and manipulates representation directions that are sensitive to modality preference, allowing for flexible and precise modulation of model behavior.

%% file: sec/dataset.tex
\section{The \textbf{MC\textsuperscript{2}} Benchmark}
In this section, we introduce the design and methodology behind the construction of the \textbf{M}ultimodal \textbf{C}ontext \textbf{C}onflict dataset, \textbf{\textsc{MC\textsuperscript{2}}}, intended for evaluating modality preference. 
We outline the data design philosophy in Section~\ref{sec:data_design}, followed by the data construction pipeline in Section~\ref{sec:pipeline} and the question design and evaluation metric in Section~\ref{sec:data_evaluate}.

\subsection{Data Design Philosophy}
\label{sec:data_design}
Modality preference is a fundamental behavioral tendency to favor one modality over another, irrespective of MLLMs’s internal knowledge and inherent understanding capabilities.
To enable a rigorous and fair assessment, we eliminate these confounding factors by ensuring that models can accurately comprehend the multi-modal context and respond based solely on the provided information. 
We elaborate on this design choice below:
1) Neutralizing Knowledge Bias: By ensuring models can derive answers directly from the context, we decouple modality preference from the interference of varying internal knowledge bases.
2) Establishing a Level Playing Field: By verifying that all models can comprehend the multi-modal context, we establish a ``common ground''—a baseline of competence that ensures a fair comparison across MLLMs with diverse capabilities.
Finally, we construct modality-conflict pairs to evaluate and compare the modality preferences of different MLLMs in a controlled environment.

\begin{figure*}[t]
 \centering
 \small
  \includegraphics[width=0.48\textwidth]{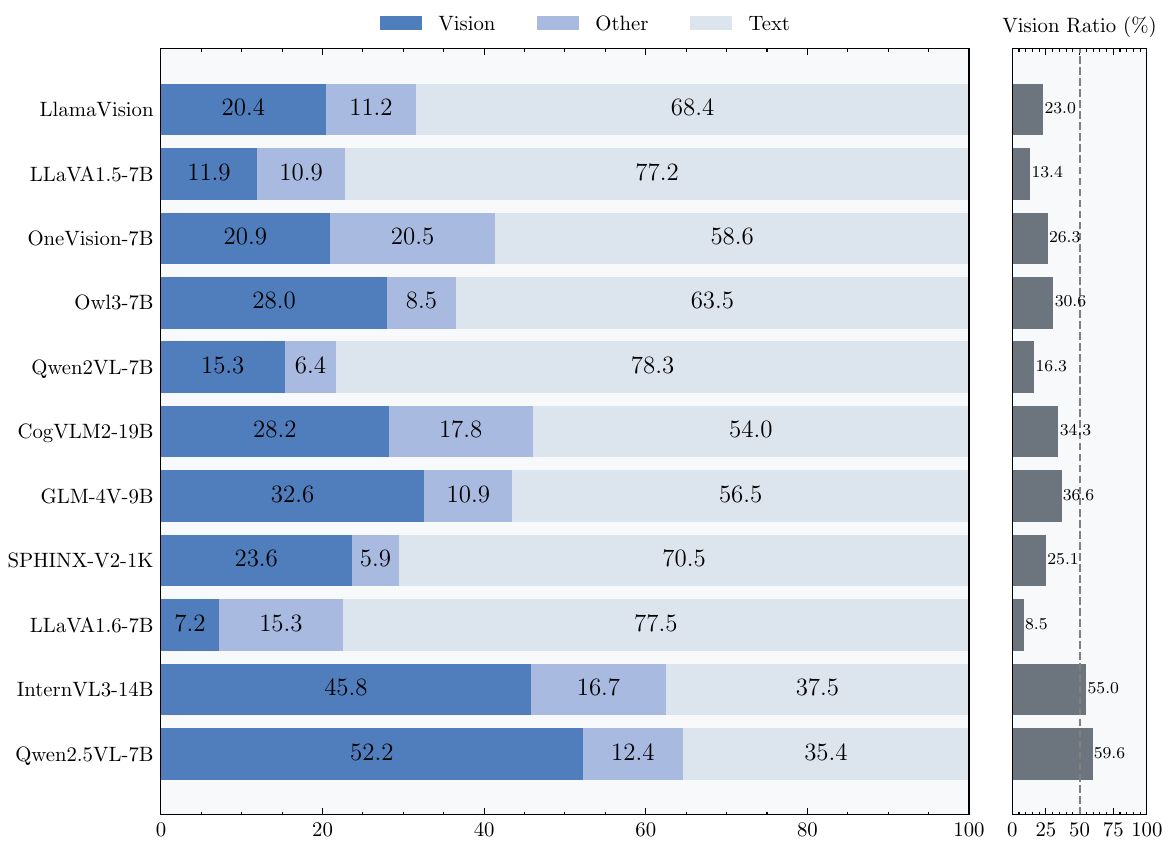} \hfill
  \includegraphics[width=0.44\textwidth]{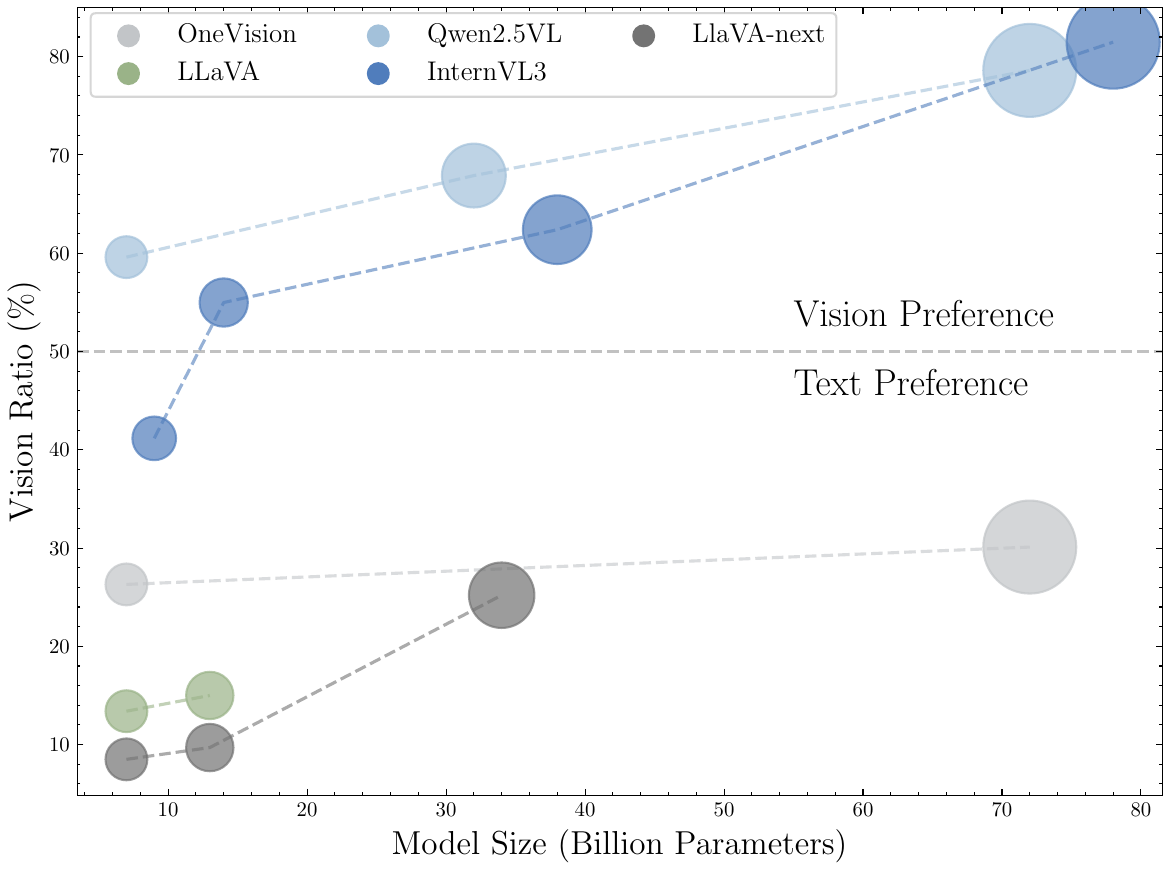} 
  \caption{Results of modality preference across different MLLMs. \textbf{Left}: Quantified scores for \textbf{Vision}, \textbf{Others} and \textbf{Text} modalities using $S_{\text{vision}}$, $S_{\text{others}}$ and $S_{\text{text}}$ as well as Vision Ratio. \textbf{Right}: Trends of Vision Ratio with respect to model parameter size.}
  \label{fig:single}
\end{figure*}
\subsection{Semi-automated Data Construction Pipeline} 
\label{sec:pipeline}

In this section, we introduce our semi-automated data construction pipeline, which follows a meticulous process to ensure the reliability of the dataset.
The data is derived from the TDIUC~\citep{tdiuc}, sourced from MS-COCO~\citep{coco}.
Therefore, the data is widely adopted in model development, ensuring that most MLLMs can recognize the images.
Firstly, we select the image as vision context $c^v$, question $q$, and answer $A^v$ based on the image and the caption $cap$ for each sample from TDIUC. 
Then the pipeline follows these steps:

\textbf{Textual Context Construction.} 
Given a sample including $c^v$, $q$, $A^v$ and $cap$, we construct candidate contrastive textual contexts $c^t$ that conflict with $c^v$ specifically in relation to $q$ but are aligned with $c^v$ and $cap$ in terms of overall scene semantics. 
We prompt DeepSeekV3~\citep{deepseek} and ChatGPT4o-mini~\citep{hurst2024gpt} to generate a distractor answer $A^t$ to $q$, together with $c^t$ that plausibly supports $A^t$, using carefully crafted instructions. 
For each model, we generate two pairs of $A^t$ and $c^t$ to facilitate downstream data selection. 
To ensure that all evaluated MLLMs can correctly recognize both visual and textual contexts, we employ several weak MLLMs, such as LLaVA1.5-7B~\citep{llava1.5} and QwenVL-7B~\citep{qwenvl}, as judges to select samples that can be correctly understood with respect to $c^v$ and $c^t$.

\textbf{Human Verification.}
We incorporate manual inspection to ensure the high quality of the data annotation. 
Specifically, we verify the existence of conflicts between $c^v$ and $c^t$ and ensure that both contexts can correctly direct $q$ to the corresponding answers, $A^v$ and $A^t$. 
Each sample is cross-verified by three human annotators to ensure the reliability of the results, and when errors are found, annotators either correct or discard the sample entirely.

\textbf{Iterative Refinement.} 
The dataset undergoes multiple rounds of refinement through a feedback loop between textual context generation and human verification, which rectifies potential errors, thus enhancing the data quality.

\textbf{\textbf{\textsc{MC\textsuperscript{2}}}.} 
To this end, we develop \textbf{\textsc{MC\textsuperscript{2}}}, a modality-conflict dataset comprising 2,000 samples spanning diverse tasks.
The detailed construction procedures for textual context generation, the manual annotation, the data format and dataset statistics are provided in Appendix~\ref{supp:data}.

\subsection{Question Design and Evaluation Metric}
\label{sec:data_evaluate}
\textbf{Question Design. }
We reformulate the original questions with ChatGPT-4o-mini~\citep{gpt4o} into a binary-choice format. 
To further reduce potential \textit{position bias} in multi-modal inputs, we adopt a \textit{consistent evaluation} strategy, similar to \citet{mmbench}. 
Concretely, for each question, we construct two versions by swapping the order of the answer choices. 
A model’s prediction is regarded as \textit{consistent} only if it selects the same answer in both versions for a sample; otherwise, it is labeled as \textit{inconsistent}. Such inconsistent samples are discarded from the subsequent measurement of modality preference.

\textbf{Evaluation Metric.}
Inspired by prior work on evaluating stylistic or knowledge-related preferences of LLMs and MLLMs through conflict-pair contexts~\citep{li2024formality,xie2023adaptive,liu-etal-2025-insight}, we extend this to evaluate modality preference by designing a metric that captures how MLLMs respond to conflicting signals from different modalities. 
More importantly, through the careful design of the benchmark, we establish as a basis that the model can reliably understand both modalities in isolation. 
As shown in Table~\ref{tab:base_text} and~\ref{tab:base_vision} in Appendix, all MLLMs achieve over 95\% accuracy, provided with either textual or visual context.

Building on this, our metric evaluates modality preference by assessing how the model’s responses align with textual or visual input when the two provide conflicting signals.
The model's response is then categorized based on which modality it aligns with: 1) \textbf{Vision}: the response aligns with the visual context; 2) \textbf{Text}: the response aligns with the textual context; 3) \textbf{Others}: the response is uncertain, or inconsistent with either modality, which is discarded from further analysis.
Then, we naturally define the \textbf{Vision Ratio} to quantify the model’s preference toward the vision modality, defined as: $ {S_{\text{vision}}}/({S_{\text{vision}} + S_{\text{text}}})$, where
$S_{\text{vision}}$ or $S_{\text{text}}$ denotes the score of the vision or text modality, computed as the proportion of samples whose responses are categorized as \textbf{Vision} or \textbf{Text} across the dataset.
$S_{\text{others}}$ is the proportion of samples whose responses are categorized as \textbf{Others}.
Vision Ratio greater than 0.5 indicates that the model tends to favor visual context over text.

%% file: sec/exp.tex
\begin{figure*}[h]
  \subfloat{\includegraphics[height=0.24\textwidth]{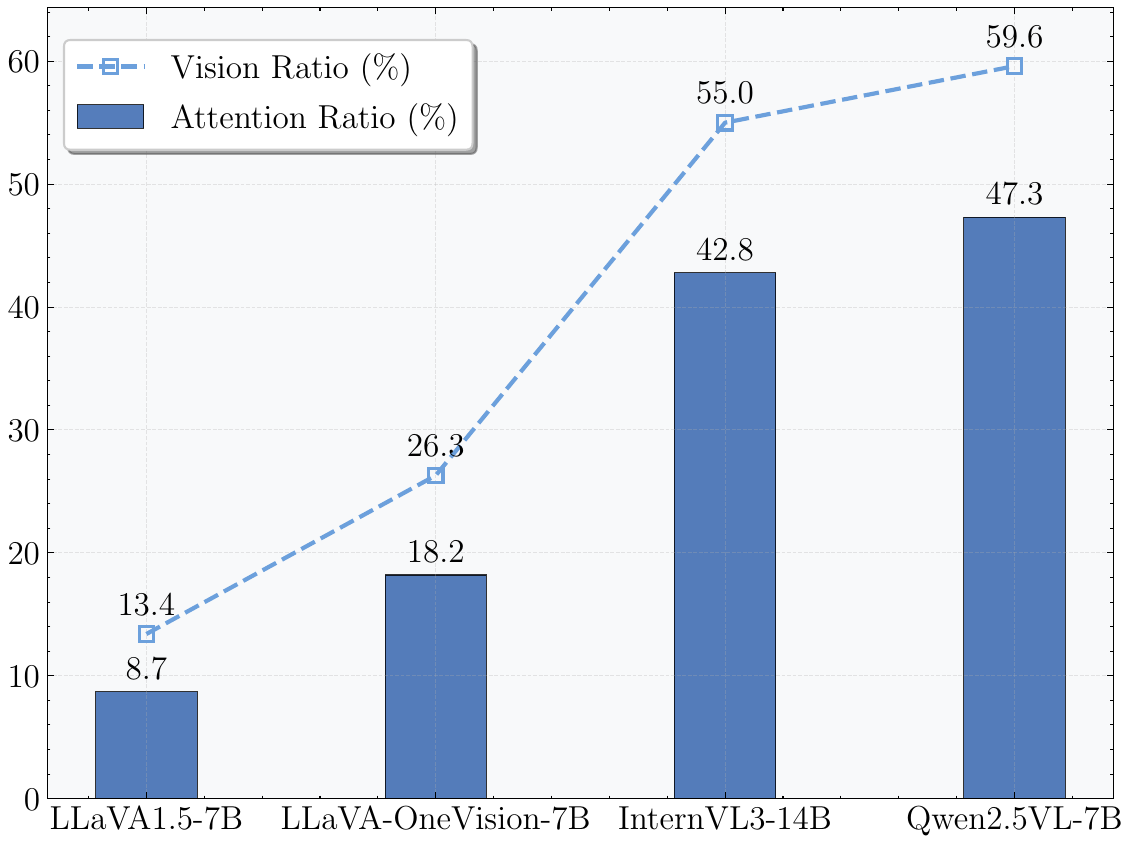}} \hfill
    \subfloat{\includegraphics[height=0.24\textwidth]{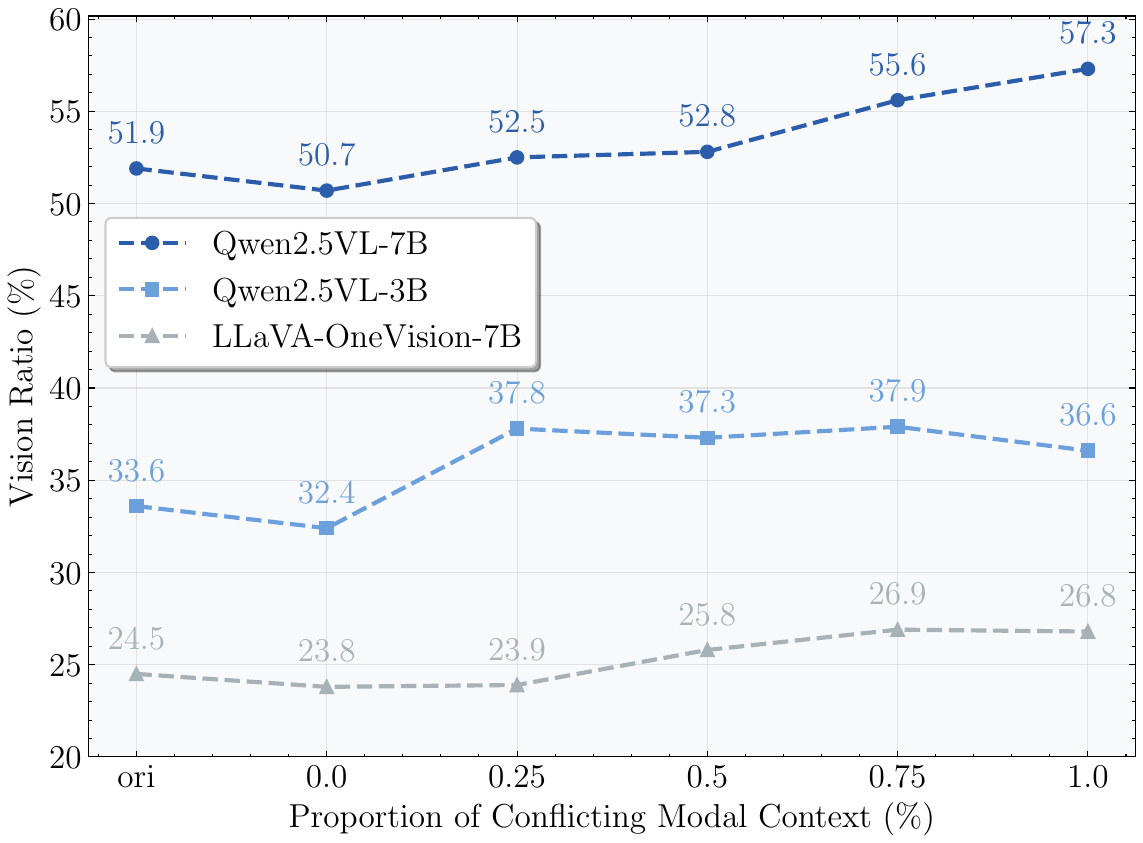}} \hfill
  \subfloat{\includegraphics[height=0.24\textwidth]{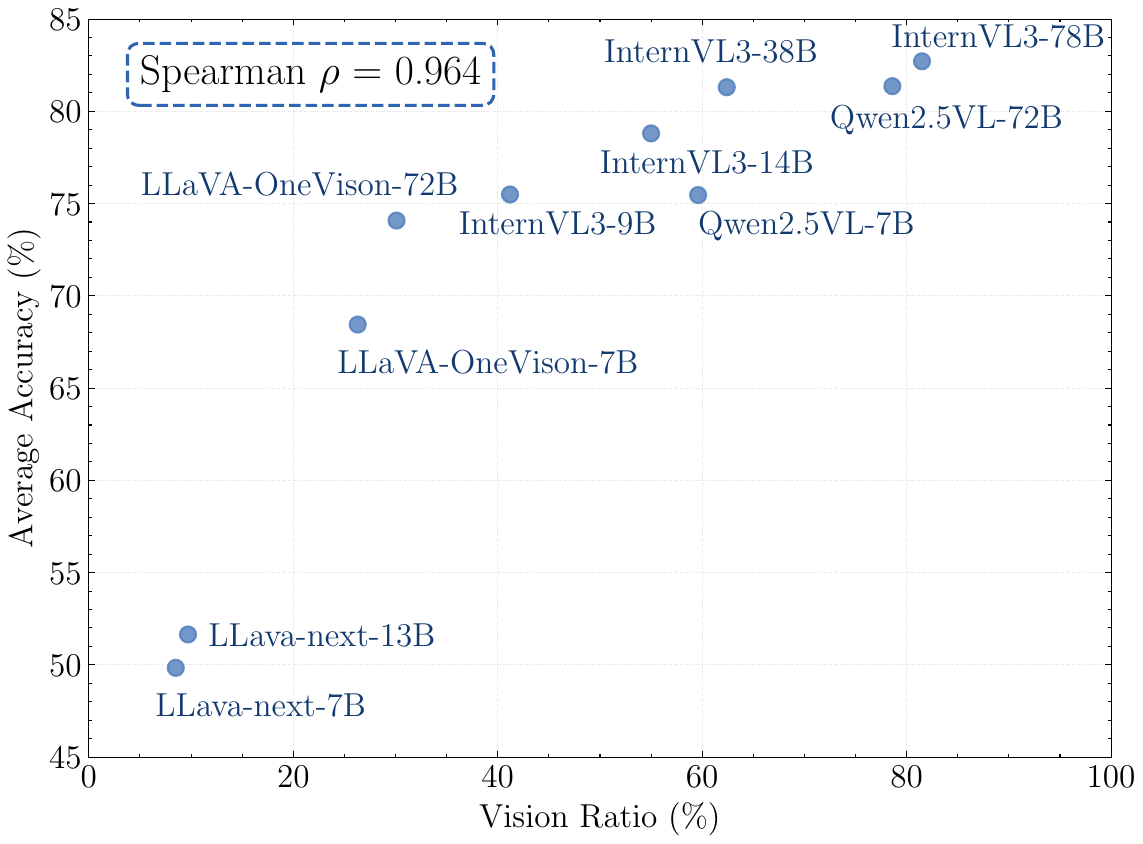}} \hfill
  \caption{Analysis of modality preferences. \textbf{Left}:  Trends of Vision Ratio and Attention Ratio. \textbf{Middle}: Vision Ratio with Respect to the proportion of multi-modal conflict-context training data and different MLLMs. \textbf{Right}: Relationship between visual understanding ability (quantified as the average accuracy across seven widely used benchmarks)  and the modality preference (measured by Vision Ratio).
}
  \label{fig:scale}
\end{figure*}
\section{Modality Preferences in MLLMs}
\label{exp}
This section presents a systematic investigation into modality preference in MLLMs, structured around four key research questions:
\textit{1) Which modality do MLLMs prioritize?} 
\textit{2) What factors drive these preferences?}
\textit{3) Can the Vision Ratio provide guidance for downstream task performance?}
\textit{4) Can modality preference be controlled?}
This investigation helps uncover the underlying mechanisms of modality preference and leverage these insights to downstream tasks.

\subsection{Which Modality Do MLLMs Prioritize?}
\label{sec:prioritize}
We use MC\textsuperscript{2} to evaluate the modality preferences of 20 open-source MLLMs and the proprietary ChatGPT-4o-mini~\citep{gpt4o}, detailed in Appendix~\ref{supp:single_modality}.

\textbf{Different MLLMs exhibit different modality preferences.}
As described in Section~\ref{sec:data_evaluate}, we quantify modality preference using  \textbf{Vision Ratio} in the left panel of Figure~\ref{fig:single} and detailed in Table~\ref{tab:single_full}.
We observe that all MLLMs exhibit clear modality preference, with most models showing a strong preference for text; for instance, LLaVA1.5-7B attains only a 13.4\% Vision Ratio.
This aligns with the previous findings that MLLMs suffer from a severe language prior~\citep{vlind,parcalabescu2024vision,wu2025foundation}.
Interestingly, Qwen2.5VL and InternVL3 show a certain degree of visual preference.

\textbf{Larger MLLMs exhibit stronger visual preferences.}
We evaluate models from the LLaVA1.5, LLaVA-Next, Qwen2.5VL, InternVL3, and LLaVA-OneVision families to investigate the relationship between model size and modality preference.
As shown in the right panel of Figure~\ref{fig:single}, we observe that for all model families, the preference for the vision modality increases with the model size.
And Qwen2.5VL and InternVL3 models exhibit a significant preference for the vision modality once the model size increases.
However, other models maintain a noticeable textual preference as model sizes increase.
To validate the reliability of the evaluation, we conduct a sensitivity analysis for the sample number of MC\textsuperscript{2} in Appendix~\ref{supp:number}.

\textbf{Vision Ratio aligns with human preference.}  
We further verify whether the Vision Ratio can serve as a human-level measure of modality preference in MLLMs. 
We randomly sample 100 instances from MC\textsuperscript{2} and compute the Vision Ratio of four representative models—Qwen2.5VL-7B, OneVision-7B, InternVL3-14B, and LLaVA1.5-7B. 
In addition, we craft prompts to elicit explicit reasoning chains from the models, specifically targeting their reliance on visual or textual information. 
To ensure labeling reliability, three expert annotators independently annotate the expressed modality preference for each response, with the final label determined by majority vote.
The automatically obtained Vision Ratio scores ([56.3\%, 24.6\%, 52.3\%, 13.9\%]) are highly consistent with those given by human annotators ([61.0\%, 22.0\%, 51.0\%, 16.0\%]), with an average discrepancy of only 2.68\%. 
This indicates that the Vision Ratio can act as a reliable, automated proxy for human assessment of modality preference.

\subsection{What Factors Drive These Preferences?}
\label{sec:reason}
Given that different MLLMs display varying modality preferences, we examine two primary sources: \textit{the internal attention distribution} and \textit{the training factors}.

\textbf{Different allocation of attention across modalities.}
We compute the mean attention scores over all token positions from both modalities and define the ratio of visual attention to total attention as the \textbf{Attention Ratio}. 
By analyzing Qwen2.5VL-7B and OneVision-7B, we observe that the trends of the Attention Ratio closely align with the Vision Ratio across models in the left panel of Figure~\ref{fig:scale}. 
This alignment suggests that MLLMs distribute attention unevenly between modalities, which in turn contributes to their divergent modality preferences.

\begin{figure*}[h]

  \subfloat{\includegraphics[height=0.24\textwidth]{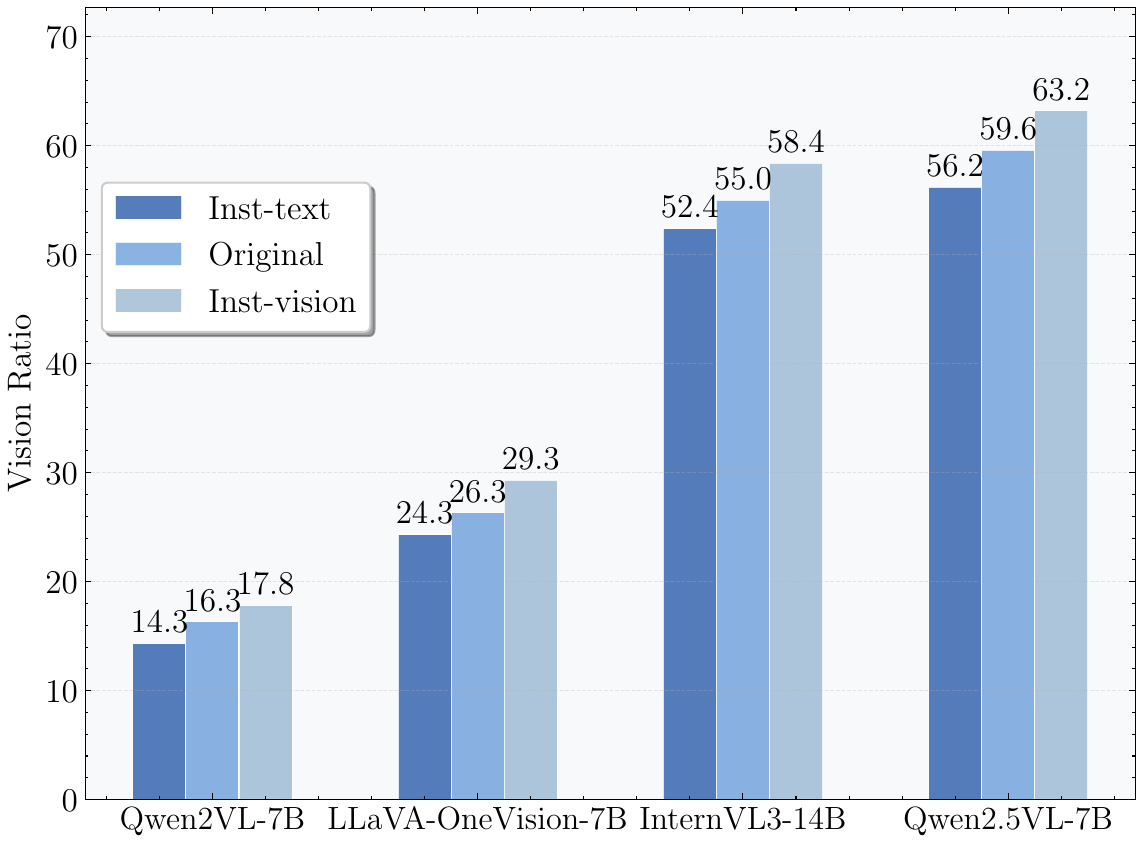}} \hfill
    \subfloat{\includegraphics[height=0.24\textwidth]{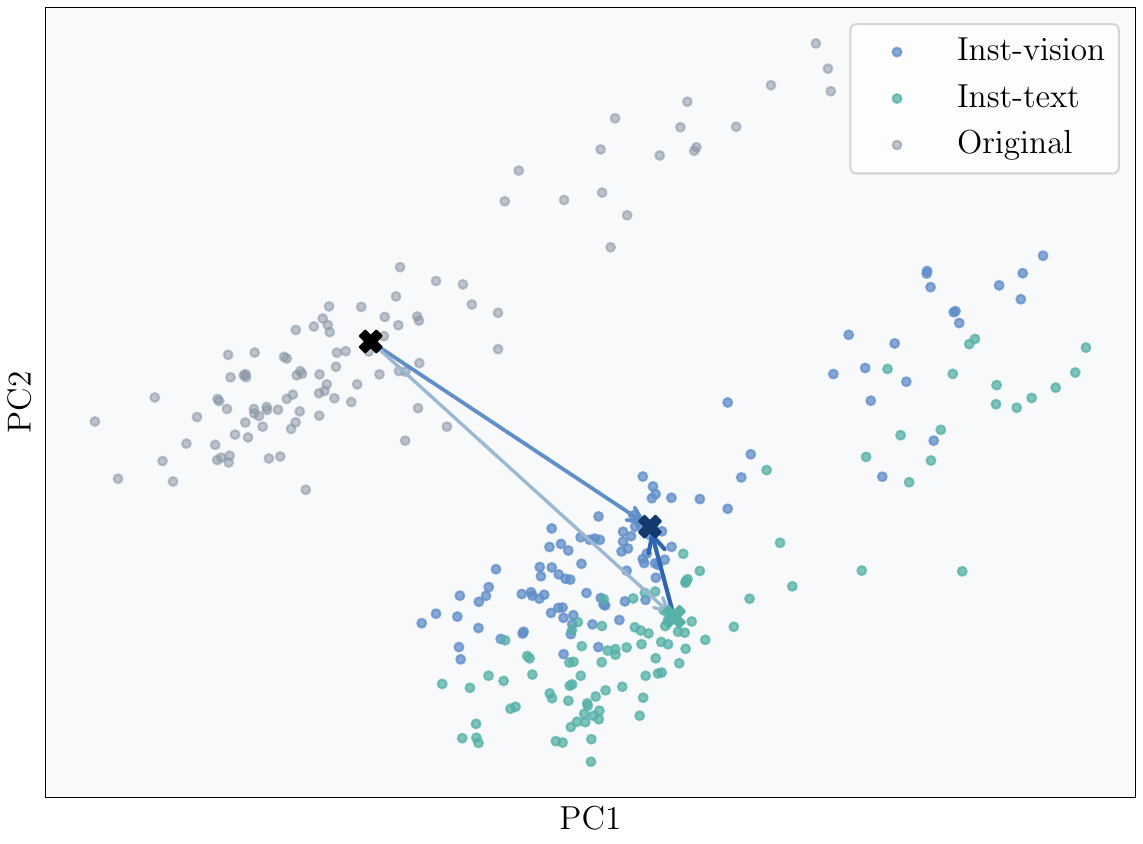}} \hfill
  \subfloat{\includegraphics[height=0.24\textwidth]{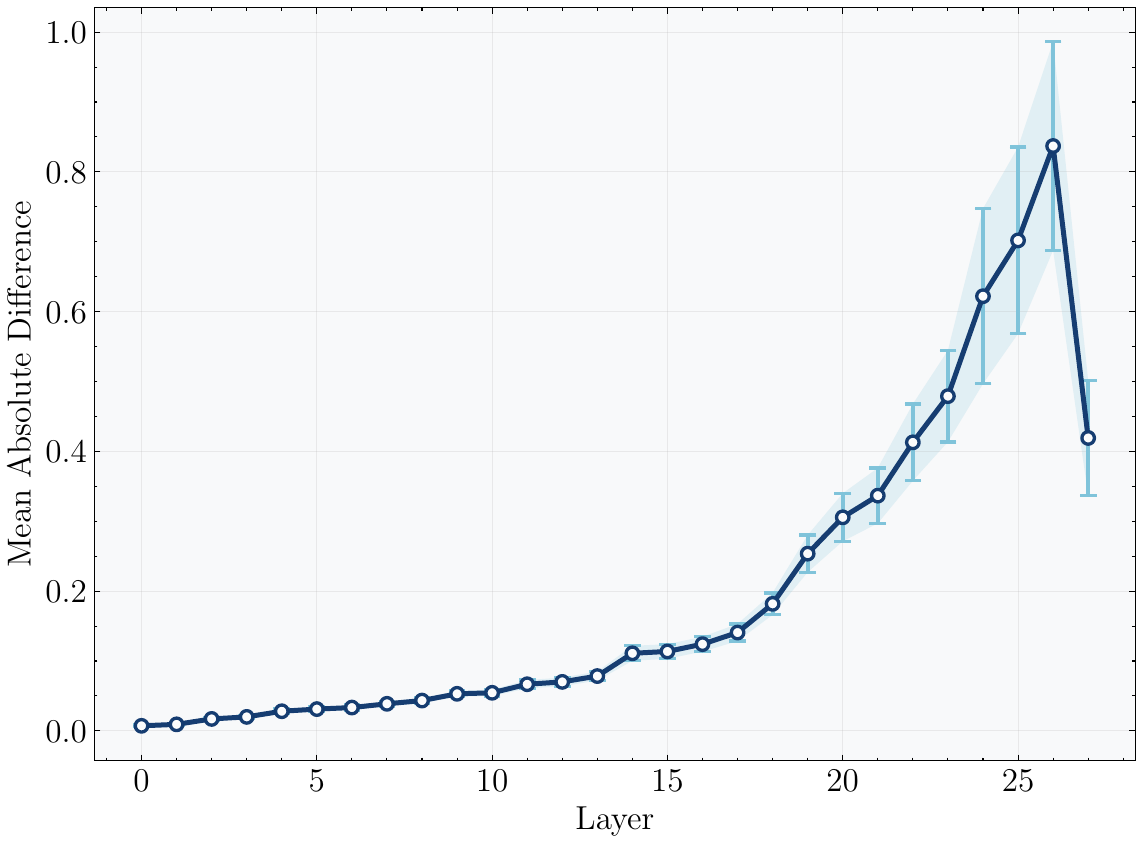}} \hfill
  \caption{
  Analysis of modality preference under instruction-guidance. \textbf{Left}:  Adjustment of modality preference using instruction-guided control (Inst-vision vs. Inst-text).
  \textbf{Middle}: Representation shifts under instruction-guided interventions.
  \textbf{Right}: Layer-wise absolute difference and standard deviation of hidden states between different instructions.
  }
  \label{fig:pca}
\end{figure*}
\textbf{Impact of model scale and training data recipe.}
Through reviewing the technical reports of the evaluated MLLMs, we find that they all adopt a common architecture comprising a vision encoder, an alignment layer, and an LLM. 
Thus, we hypothesize that the observed preferences mainly arise from two factors: 
 1) Exposure to more multi-modal contexts, especially with conflicting cases, drives more pronounced shifts in modality preference;
 2) Larger LLMs are more capable of shifting their preference during training.

To examine these hypotheses, we construct a training dataset containing vision–text conflict contexts and fine-tune Qwen2.5VL-7B/3B and LLaVA1.5-7B with varying proportions of samples with multi-modal conflict contexts to adjust preferences toward text or vision. 
We optimize MLLMs in the opposite direction of their original preferences and measure changes using the Vision Ratio. 
As shown in the middle panel of Figure~\ref{fig:scale}, increasing the proportion of multi-modal contexts consistently leads to larger preference shifts, supporting \textbf{Hypothesis 1}. 
This suggests that multi-modal inputs may create more challenging conditions, leading to greater preference shifts.
Furthermore, Qwen2.5VL-7B exhibits greater shifts than Qwen2.5VL-3B under the same training data, supporting \textbf{Hypothesis 2}.
This indicates that  larger LLMs demonstrate stronger adaptability.

\subsection{Can the Vision Ratio provide guidance for downstream task performance?}
\label{sec:vision_ratio_dis_post}
As a foundational behavioral prior, the identified modality preference can inform how a model integrates information across modalities. As such, the findings can offer relevant insights into the model’s behavior in deeper cross-modal understanding tasks.
To demonstrate the correlation between the modality preference and performance on downstream tasks, we evaluate the visual understanding abilities of 10 representative MLLMs. 
Specifically, we compute the average accuracy across 7 widely used benchmarks, as detailed in Appendix~\ref{supp:vision_ratio_dis_post}.
We then compare the visual understanding abilities with their modality preference measured by Vision Ratio using MC\textsuperscript{2}, as shown in the right panel of Figure~\ref{fig:scale}.  
The results reveal a strong positive association between Vision Ratio and visual understanding ability across all MLLMs. 
Notably, Spearman’s rank correlation~\citep{sedgwick2014spearman} reaches $\rho = 0.964$, demonstrating that Vision Ratio provides a highly reliable indicator of visual understanding ability.

\subsection{Can Modality Preference Be Controlled?}
\label{sec:interventions}
We investigate whether modality preference can be controlled, and conduct latent space representation analysis to examine the underlying mechanisms.
The detailed setting and results are provided in Appendix~\ref{supp:control}.

\textbf{Modality preference can be guided through instruction design.}
We investigate the impact of instruction design on modality preference, for instance, by explicitly directing the model to rely on a specific modality when answering a question. 
Specifically, we evaluate the modality preference using Vision Ratio for four representative MLLMs including Qwen2.5VL-7B, OneVision-7B, Qwen2VL-7B and InternVL3-14B under the text and vision preference instruction (Inst-text and Inst-vision).
As shown in the left panel of Figure~\ref{fig:pca}, instructions, probing the models toward any modality effectively shape their modality preferences.

\textbf{Modality preference direction in representation space.}
To further understand how the intervention methods influence modality preference internally, we analyze the hidden representations of the models. 
Specifically, we apply Principal Component Analysis (PCA;~\citealp{pca}) to the hidden states to identify the dominant direction of modality preference. 
The middle panel of Figure~\ref{fig:pca} shows that instruction-based interventions drive clear shifts in representations, revealing that the model’s internal states are sensitive to modality control. This motivates us to develop specific techniques to adjust modality preference in Section~\ref{sec:method}.

%% file: sec/method.tex
\begin{figure*}[t]
    \centering
    \centerline{\includegraphics[width=0.9\linewidth]{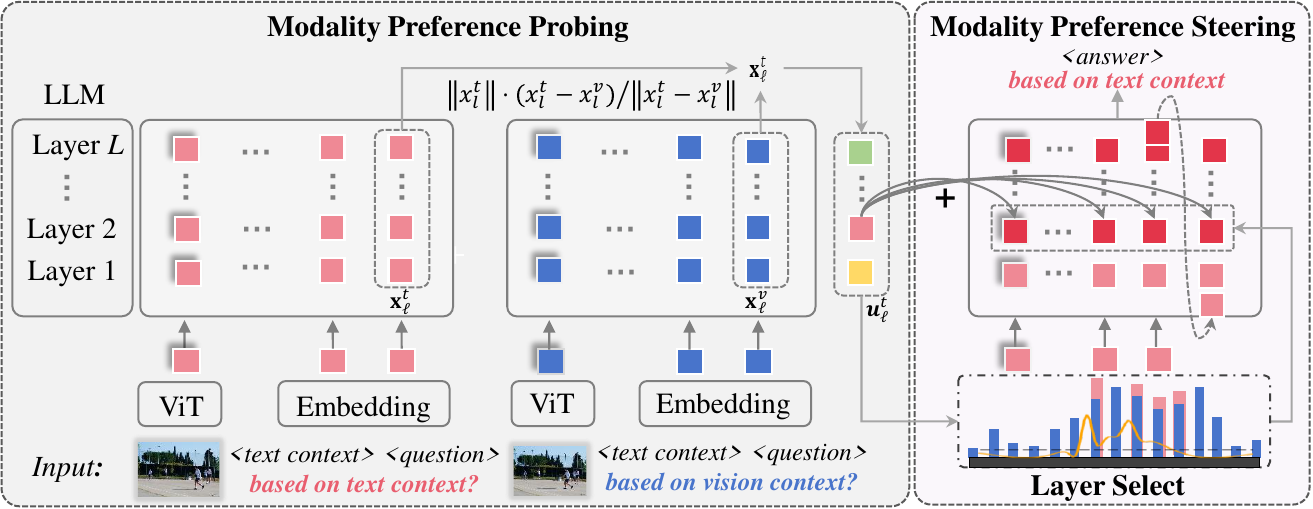}}
    \caption{Overall framework of the proposed method.
    Modality Preference Probing collects the neural activity, computes and scales the direction of modality preference.
    Modality Preference Steering selects the target layer during the second inference and adds the scaled modality preference direction to the representation at the corresponding layer at each inference step.}
    \label{fig:method}
\end{figure*}
\section{Method}
\label{sec:method}
Inspired by the behavior discussed in Section~\ref{sec:interventions}, we propose to use representation engineering~\citep{zou2023representation} to steer modality preference.
As shown in Figure~\ref{fig:method}, the proposed framework consists of Modality Preference Probing (\S\ref{sec:MPP}) and Modality Preference Steering (\S\ref{sec:MPS}).

\subsection{Modality Preference Probing}
\label{sec:MPP}

We probe and collect neural activity that represents the direction of modality preference. 
Inspired by the pre-training Next Token Prediction objective of decoder-only MLLMs~\citep{gpt4o} and the method to extract classification features~\citep{lastToken}, we collect neural activity from the last token in the input text.
The process involves probing modality preference through two requests: one with a vision preference probing (e.g., `answer the question based on the vision context') and another with a text preference probing (e.g., `answer the question based on the text context').
Let us denote these inputs by $q^v$ and $q^t$, and consider a set of $N$ such pairs ($q^v_i, q^t_i$), $i \in \{1,\dots,N\}$.
Let $\mathbf{x}_{i,\ell}^v, \mathbf{x}_{i,\ell}^t \in \mathbb{R}^d$ be the hidden states for the two queries at the last token at layer $\ell \in \{1,\dots,L\}$, where $d$ is the dimension of the chosen MLLM.
We identify the direction of modality preference by computing the difference in the hidden states between the paired inputs. 
Formally, we compute a vector $\mathbf{u}_\ell \in \mathbb{R}^{d}$ representing the direction towards the text modality at layer $\ell$ for a given query as:
\begin{align}
\label{eq:probing}
    \mathbf{u}^t_\ell = \frac{1}{N}\sum_{i}^{N} \big(\mathbf{x}_{i,\ell}^t - \mathbf{x}_{i,\ell}^v \big).
\end{align}
\begin{table}[t]
\centering
\caption{Performance for steering Qwen2VL-7B and OneVision-7B towards vision modality, and Qwen2.5VL-7B and InternVL3-8B towards text modality, measured by $S_{\text{vision}}$ and $S_{\text{text}}$.}
\resizebox{\linewidth}{!}{

\begin{tabular}{llccccc}
\toprule
\textbf{Preference} & \textbf{Model}
 & \textbf{MLLM} & \textbf{InstDesign} & \textbf{CoT} & \textbf{FewShot} & \textbf{Ours} \\
\midrule
\multirow{2}{*}{\textbf{Text}$\uparrow$}  
 & \textbf{Qwen2.5VL-7B}   & 35.4 & 37.7 & 55.6 & 61.1 & \textbf{63.6 }\\
 & \textbf{InternVL3-8B} & 20.9   & 31.6   & 36.7   & 38.2   & \textbf{42.8}   \\
\midrule
\multirow{2}{*}{\textbf{Vision}$\uparrow$} 
 & \textbf{Qwen2VL-7B} & 15.3 & 32.3 & 34.2 & 17.2 & \textbf{48.1}\\
 & \textbf{OneVision-7B} & 37.5   & 52.8  & 53.1   & 49.8   & \textbf{57.1}   \\
\bottomrule
\end{tabular}
}

\label{tab:model_performance_transposed}
\end{table}

Averaging over different queries allows us to capture the activation values most closely associated with modality preference, independent of questions.
As shown in the right panel of Figure~\ref{fig:pca}, we compute the absolute values and standard deviations of the modality preference direction $\mathbf{u}^t_\ell$ across different samples. 
We observe that layers 20–23 exhibit higher absolute values and lower variance, indicating that the preference direction is more prominent and stable in these layers. 
Based on this observation, we select the corresponding layer $\ell'$ of the model to control the direction of modality preference in Section~\ref{sec:MPS}.
Similar patterns are observed for Qwen2VL-7B, Qwen2.5VL-7B, LLaVA-OneVision and InternVL3, as detailed in Appendix~\ref{sec:pattern}.

\begin{table}[t]
    \centering
    \caption{Performance of the proposed method on the reasoning tasks (MathVista, TallyQA and VSR) measured by accuracy for Qwen2.5VL-7B and multi-modal translation task (Ambigcaps) measured by BLEU scores for English (En) $\leftrightarrow$ Turkish (Tr).}
    \label{tab:reasoning}
    \begingroup %
    \scalebox{0.8}{
    \begin{tabular}{l ccc}
        \toprule
        \textbf{Dataset} & \textbf{CoT} & \textbf{InstDesign} & \textbf{Ours} \\
        \midrule
        MathVista & 50.0 & 59.0 & \textbf{62.3} \\
        TallyQA & 61.6 & 74.4 & \textbf{76.6} \\
        VSR & 45.6 & 51.3 & \textbf{53.2} \\
        Ambigcaps (\textbf{En\textrightarrow Tr}) & 9.03 & 9.45 &\textbf{10.2} \\
        Ambigcaps (\textbf{Tr\textrightarrow En}) & 18.63 & 18.98 & \textbf{19.89} \\
        \bottomrule
    \end{tabular}
    }
    \endgroup 
\end{table}

\begin{table}[t]
\centering
\caption{Performance of the proposed method on the visual understanding benchmark, Phd~\citep{phd}. we report the accuracy results on the phd-icc/phd-iac.}
\renewcommand\arraystretch{1.}
\scalebox{0.6}{
\begin{tabular}{lcccccc}
\toprule
\textbf{Model}  & \textbf{Attribute} & \textbf{Sentiment} & \textbf{Positional} & \textbf{Counting}  & \textbf{Object} & \textbf{Avg} \\
\midrule
\textbf{Qwen2VL-7B}
& 10.0 / 28.5 & \enspace2.5 / 8.5  & 3.5 / 20.5 & 6.0 / 30.5 & \enspace8.0 / 50.0  & \enspace6.0 / 27.6  \\
\textbf{+InstDesign} 
& 14.5 / 34.5 & \enspace2.5 / 13.0 & 1.5 / 26.0 & 5.5 / 39.0 & 25.0 / 60.0 & \enspace9.8 / 34.5  \\
\textbf{+CoT} 
& \enspace5.0 / 15.5  & \enspace6.0 / 23.5 & 8.5 / 30.2 & 6.5 / 17.0 & 40.5 / 59.0 & 13.3 / 29.0 \\
\textbf{+FewShot} 
& \enspace3.0 / 17.0  & \enspace0.5 /\enspace9.0  & 1.5 / 14.5 & 5.0 / 29.0 & \enspace2.0 / 37.0  & \enspace2.4 / 21.3  \\
\textbf{+Ours} 

& 10.0 / 34.4 & 11.0 /16.5 & 14.0 / 28.3& 5.0 / 37.4 & 51.5 / 64.0 & \textbf{18.4 }/ \textbf{36.1} \\
\midrule

\textbf{OneVision-7B} 
&  11.5/ 20.5 & \enspace1.5 / 5.0  & \enspace1.5 / 16.5 &  \enspace6.5 / 28.5  & 11.0 / 52.0 & \enspace6.4 / 24.5  \\
\textbf{+InstDesign}
& 16.0 / 27.0 & \enspace5.5 / 12.5 & \enspace6.0 / 31.5 & 13.5 / 30.5 & 34.0 / 61.5 & 15.0 / 32.6 \\
\textbf{+CoT} 
& 17.3 / 28.4 & \enspace6.2 / 12.9 & \enspace7.8 / 33.2 & 13.8 / 30.9 & 34.5 / 62.1 & 15.9 / 33.1 \\
\textbf{+FewShot} 
& 17.0 / 28.0 & \enspace6.0 / 13.0 & \enspace7.2 / 32.8 & 13.9 / 31.0 & 34.8 / 62.3 & 16.2 / 33.4 \\
\textbf{+Ours} 
& 19.6 / 30.5 & \enspace7.8 / 13.5 & 10.3 / 36.4 & 15.1 / 29.8 & 35.6 / 63.5 & \textbf{17.7} / \textbf{34.7} \\
\bottomrule
\end{tabular}
}
\label{tab:phd_main}
\end{table}

\subsection{Modality Preference Steering}

\begin{figure*}[t]
  \subfloat{\includegraphics[height=0.24\textwidth]{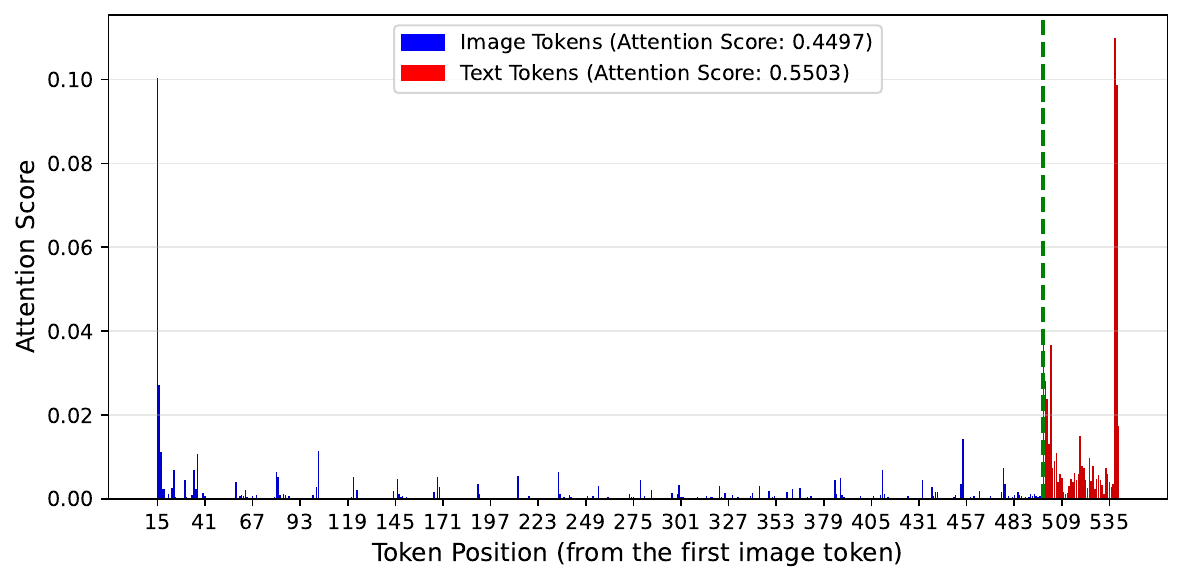}} \hfill
  \subfloat{\includegraphics[height=0.24\textwidth]
  {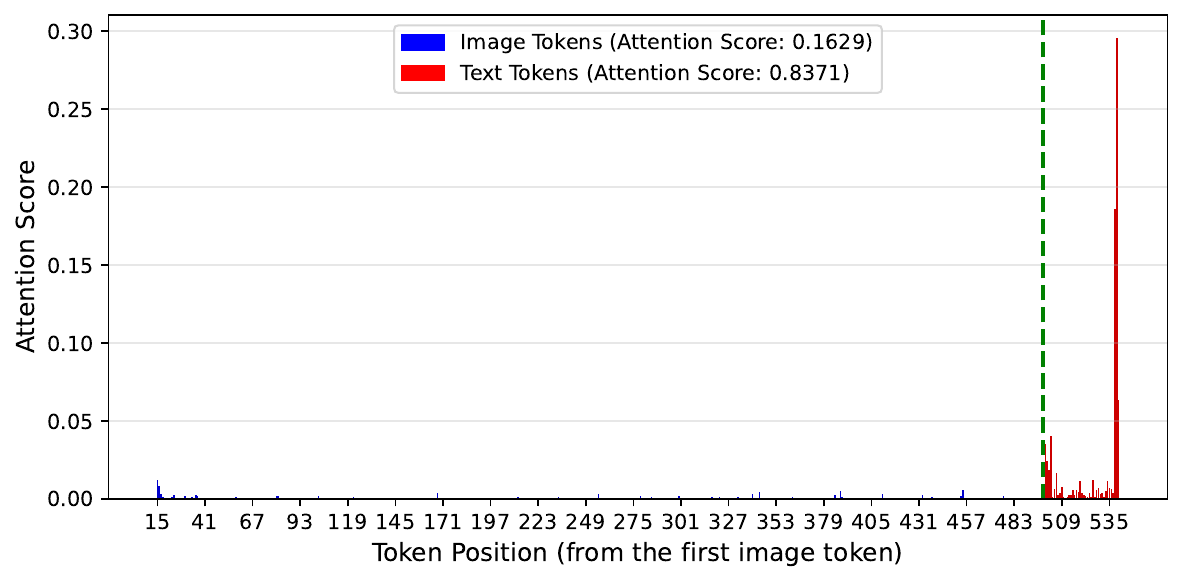}} \hfill
  \caption {The attention scores of the generated token toward the vision and text contexts at the 22th layer using InstDesign (Left) and
  the proposed method (Right).}

  \label{fig:attention_steer}
\end{figure*}
\label{sec:MPS}
After obtaining the probing direction vector, we compute the steering vector by re-scaling the vector $\mathbf{u}^t_\ell$ with a weight $w$.
The scaling process must carefully balance two objectives: 1) it must be strong enough to effectively steer the model's modality preference, 2) it must preserve the model’s normal output behavior.
In our preliminary experiments, we observe that setting the weight too large leads to repetitive and meaningless outputs, whereas a too small weight fails to obtain any noticeable change for modality preference.
Unlike previous approaches~\citep{zou2023representation,stolfo2024improving} that rely on exhaustive search over a validation set to determine the weight, we propose a principled method that aligns the mean of the probed direction distribution with the mean of the original distribution of hidden states.
This strategy ensures that the steering remains effective without disrupting the model's inherent generation capabilities.
Formally, the weight is determined by aligning the mean of the probed direction with the central distribution of the original hidden states and the steering vector is computed by:
\begin{align}
\label{eq:steering}
    \mathbf{s}^t_\ell=w\mathbf{u}^t_\ell,\ \text{where}\ w=  \frac{1}{N} \sum_{i=1}^{N}  \frac{\| \mathbf{x}^t_{i,\ell} \|}{\| \mathbf{u}^t_\ell \|}     
\end{align}

Finally, during the second inference, we adjust the hidden states at the selected layer $\ell'$ at each decoding step by adding $\mathbf{s}^t_{\ell'}$ to all tokens to steer the model's response toward text modality.
Similarly, steering towards vision modality is performed towards the opposite direction and no additional data or labels are introduced.

\subsection{Experiments} 
We evaluate the proposed method for controlling modality preference on \textbf{\textsc{MC\textsuperscript{2}}} and multiple downstream tasks across four representative MLLMs: Qwen2VL-7B, Qwen2.5VL-7B, OneVision-7B, and InternVL3-8B. We compare our approach against several widely-used training-free baselines: \textit{MLLM} (zero-shot direct response), \textit{InstDesign} (instruction-based guidance for modality preference), \textit{CoT} (utilizing intermediate processing steps), and \textit{FewShot} (four-shot in-context learning). Implementation details are provided in Appendix~\ref{supp:applying}.
As shown in Table~\ref{tab:model_performance_transposed}, our method consistently outperforms all baselines on \textbf{\textsc{MC\textsuperscript{2}}} across both evaluation settings.
These results demonstrate the superior effectiveness of our approach in precisely adjusting modality preference compared to standard prompting techniques.

We further assess our method across a variety of downstream scenarios:
1) Multi-modal reasoning tasks on MathVista~\cite{mathvista}, TallyQA~\cite{tallyqa} and VSR~\cite{vsr}; 2) Multi-modal understanding tasks on PhD~\citep{phd} and AmbigCaps~\citep{ambigcaps}, a multi-modal machine translation (MMT) task.
The Phd dataset includes two subsets—PhD-ica, which contains irrelevant textual context, and PhD-icc, which introduces misleading or incorrect textual information—both of which increase the risk of hallucination.

As shown in Table~\ref{tab:reasoning}, our method consistently surpasses both the \textit{CoT} and \textit{InstDesign} baselines. 
We observe that by steering the model toward a visual preference, our method increases reliance on the original visual input. 
This effectively prevents the final decision from being misled by potential vision hallucination that often emerges during the sequential steps of CoT.
On the AmbigCaps translation task, our method achieves a significant improvement of $1.33$ BLEU score over the strongest baseline.
Furthermore, Table~\ref{tab:phd_main} demonstrates substantial improvements in multi-modal understanding. Specifically, when applied to Qwen2VL-7B, our method surpasses the best-performing baseline by an average margin of $6.1$ percentage points. 
Additional results for more tasks and models are available in Appendix~\ref{supp:more_results_}.

\subsection{In-depth Analysis for Steering Method}
\label{sec:in_depth}
To investigate the internal mechanism of the steering method, we analyze the attention scores of the generated token toward the vision and text contexts using the proposed steering method and the InstDesign method with the samples from $MC^2$. 
In Figure~\ref{fig:attention_steer}, we visualize the attention distribution at the  24th layer by steering the Qwen2.5VL-7B at the 22nd layer towards the text modality using the case in Figure~\ref{fig:attention_detail}, as detailed in Appendix~\ref{supp:details_analysis}.
We observe that after applying the steering method, the model's attention weight toward the text modality significantly increases. This change clearly demonstrates that the steering mechanism successfully alters the modality preference by enhancing the model's dependency on the text modality.
More ablation experiments, the analysis of the latency, memory usage and the prerequisites for the proposed method are detailed in Appendix~\ref{supp:ablation} and ~\ref{supp:details_analysis}.

%% file: sec/conclusion.tex
\section{Conclusion}

This paper investigates modality preference in multi-modal large language models (MLLMs).
We carefully curate a modality conflict dataset and use a controlled experimental setup to quantitatively evaluate modality preference. 
Besides, we find the direction of modality preference can be captured within the latent representations of MLLMs.
Inspired by this, we propose a modality preference probing and steering method, which enables significant and flexible changes in modality preference. 
Experiments show that the proposed method generalizes well to downstream tasks, such as multi-modal reasoning and understanding tasks.

%% file: sec/supp.tex
\newpage
\appendix
\onecolumn

All codes, data, and instructions for our MC\textsuperscript{2} can be found in \url{https://github.com/EchoDreamer/Modality-Preference}. MC\textsuperscript{2} is released under a Creative Commons Attribution 4.0 License (CC BY 4.0).

Our supplementary materials are summarized as follows:
\begin{itemize} 
    \item Discussion: Limitations, Strategic Insights for Future Research, Use of LLM and License of Assets.
    
    \item Dataset Construction
    
    \item Model Evaluation
    
    \item Method Applying
    \item More Experiment Analysis
\end{itemize}

\section{Discussion}
\label{supp:disc}

\subsection{Limitations}
This paper investigates the modality preference in multi-modal large language models (MLLMs) using a controlled experiment setup with a modality conflict dataset.
In constructing the dataset, we employs LLaVA1.5-7B and QwenVL-7B to filter samples and ensure that most models could answer questions correctly based on a single modality. 
However, this process requires multiple iterations and turned out to be time-consuming. 
Therefore, devising a more efficient and elegant method for sample selection may be of greater importance.

\subsection{Strategic Insights for Future Research}
Based on the experiment analysis, we highlight two key principles for designing effective multi-modal learning strategies:
1) As observed in Section~\ref{sec:reason}, increasing the proportion of data with  multi-modal context (\textbf{TV}QA)  accelerates the adjustment of modality preference. Therefore, for MLLMs—especially those inheriting parameters from LLMs—introducing a more TVQA data can more rapidly mitigate the inherent language bias, facilitating faster convergence toward balanced multi-modal learning.

2)  We identify that the Vision Ratio serves as a reliable indicator of visual understanding, and adjusting modality preference improves downstream performance. Consequently, combined with the first point, selecting an appropriate ratio of TVQA training data can adjust modality preferences to satisfy the specific requirements of different downstream.

\subsection{Use of LLM}
In this work, we use the LLMs including GPT-4o-mini~\citep{gpt4o} and DeepSeekV3~\citep{deepseek}, and MLLMs including LLaVA1.5-7B~\citep{llava1.5} and QwenVL-7B~\citep{qwenvl} to help annotate the text context in MC\textsuperscript{2}, as detailed in Section~\ref{supp:data}.
We evaluate the modality preference of 20 open-source MLLMs and GPT-4o-mini~\citep{gpt4o}, and steer the modality preference of Qwen2VL-7B~\citep{qwen2vl}, Qwen2.5VL-7B~\citep{qwen2.5vl}, and OneVision-7B~\citep{onevision} and InternVL3-8B~\citep{chen2024internvl}.
Besides, we also utilize the LLMs to correct the grammatical errors.

\subsection{License of Assets}
All images in  MC\textsuperscript{2} are publicly available from COCO~\citep{coco}.
We release our benchmark under a Creative Commons Attribution 4.0 License (CC BY 4.0) to enhance global accessibility and foster innovation and collaboration in research.

\section{Dataset Construction}
\label{supp:data}
\subsection{Conflict Text Context Generation}
\label{supp:prompt}
\paragraph{Details for data generation using LLMs}
To ensure reproducibility and transparency, we include the exact prompts used in our data generation process. 
These prompts were designed to generate the candidate textual contexts and corresponding answers using GPT-4o-mini~\citep{gpt4o} and DeepSeekV3~\citep{deepseek}.
Below, we provide representative examples of the prompts used during dataset construction given the caption of an image, question, the answer for the question based on image and the task type for the question. 
For the full list of prompts, please refer to the project repository.

\begin{tcolorbox}[colback=gray!5!white,colframe=black!75!white,title=Conflict Context Generation for counting task using DeepSeekV3]
Instruction: \\
\# Given a description of an image and a corresponding counting type question with its answer, now you are required to generate a text context that points to an answer that fluctuates by 1 or 2 from the original answer. The context explicitly supports the new answer, providing clear evidence that aligns logically with the counting question. Only one alternative answer should be generated. \\
Caption: \{\textbf{caption}\} \\
Question: \{\textbf{question}\} \\
Answer: \{\textbf{answer based on vision context}\} \\
Output the new answer enclosed in <answer> </answer> and the context enclosed in <context> </context> tags. 
\end{tcolorbox}

\begin{tcolorbox}[colback=gray!5!white,colframe=black!75!white,title=Conflict Context Generation of for other tasks using DeepSeekV3]
Instruction: \\
\# Given the caption of an image and a corresponding \{\textbf{task-type}\} type question with its answer, now you are required to generate a text context as a premise that supports a new distractor answer for the question. The context should mimic the environment described in the caption but should not include \{\textbf{answer based on vision context}\}, while maintaining logical consistency within the context. Only one alternative answer should be generated. \\
Caption: \{\textbf{caption}\} \\
Question: \{\textbf{question}\} \\
Output the new answer enclosed in <answer> </answer> and the context enclosed in <context> </context> tags. 
\end{tcolorbox}

\begin{tcolorbox}[colback=gray!5!white,colframe=black!75!white,title=Conflict Context Generation for other tasks using GPT-4o-mini]
Instruction: \\
\# Given a caption of an image and a corresponding counting question with its answer, you are required to generate a single text context that provides an indirect premise leading to a new answer that fluctuates by 1 or 2 from the original answer. The context should build an indirect premise to the new answer. Carefully design this context. For this task, I want you to first describe the scene with a certain quantity and then introduce an increase or decrease in that quantity to imply the final answer and don't include the final answer. Only one alternative answer should be generated. \\
Caption: \{\textbf{caption}\} \\
Question: \{\textbf{question}\} \\
Answer: \{\textbf{answer based on vision context}\} \\
Task-type: \{\textbf{task-type}\} \\
Output the new answer enclosed in <answer> </answer> and the context enclosed in <context> </context> tags. 
\end{tcolorbox}

\begin{tcolorbox}[colback=gray!5!white,colframe=black!75!white,title=Conflict Context Generation for count task using GPT-4o-mini]
Instruction: \\
\# Given the caption of an image and a corresponding question with its answer, now you are required to generate a text context as the indirect premise of a new answer for the question, which belongs to the same category as the original answer. The context should support the new answer, include the caption while maintaining logical consistency within the context and don't include the final answer.  Only one alternative answer should be generated. \\
Caption: \{\textbf{caption}\} \\
Question: \{\textbf{question}\} \\
Answer: \{\textbf{answer}\} \\
Task-type: \{\textbf{task-type}\} \\
Output the new answer enclosed in <answer> </answer> and the context enclosed in <context> </context> tags. 
\end{tcolorbox}

\paragraph{Human Verification}
\label{supp:human}
Although the text contexts and answers generated by strong LLMs—filtered through judge MLLMs such as LLaVA1.5-7B~\citep{llava1.5} and QwenVL-7B~\citep{qwenvl}—generally yield reliable results, we further incorporate manual inspection to ensure the high quality of data annotations.
Specifically, we verify that the visual and textual contexts are indeed in conflict, and that each modality independently supports the corresponding answer to the given question. 
This involves a two-stage manual review process:

\begin{itemize}
    \item \textbf{Modality-Answer Alignment.} First, for each context from different modalities (image and text), annotators assess whether it independently provides sufficient information to correctly answer the question. 
    This step is particularly important because the original VQA answers in the TDIUC~\citep{tdiuc} dataset may contain error annotations, and the LLM-generated contexts and answers may occasionally be inconsistent.
    \item \textbf{Conflict Verification.} Next, annotators examine whether the visual and textual contexts are semantically inconsistent with respect to the question. 
    That is, the two modalities should lead to different correct answers when considered separately. Samples where both modalities lead to the same answer are discarded, as they do not reflect a true modality conflict.
\end{itemize}

Samples that do not meet either verification criterion are flagged for further review. 
Depending on the nature and severity of the issue, we take one of the following actions: revise the prompt to improve clarity, regenerate the problematic part of the sample (e.g., the question or context), or discard the sample entirely if it cannot be reasonably corrected.

To ensure consistency and reduce subjectivity, each category (i.e., vision-aligned, text-aligned, and conflict) is independently verified by three trained annotators. 
Disagreements are resolved through discussion or majority voting. 
In addition, we conduct random spot-checks throughout the dataset to ensure the consistency and reliability of the annotations.
\begin{table}[h]
\centering
\caption{Average text context length across different task types in the MC\textsuperscript{2} dataset.}
\resizebox{\linewidth}{!}{
\renewcommand\arraystretch{1.2}
\begin{tabular}{lccccccccc}
\toprule
\textbf{statistics} & \textbf{Sport} & \textbf{Attribute} & \textbf{Sentiment} & \textbf{Positional} & \textbf{Counting} & \textbf{Color} & \textbf{Activity} & \textbf{Object} & \textbf{Avg} \\
\midrule
\textbf{Text Length} & 52.48 & 33.50 & 39.69 & 31.53 & 37.12 & 31.15 & 49.68 & 39.71 & 39.36 \\

\bottomrule
\end{tabular}
}

\label{tab:dataset_statistics}
\end{table}
\subsection{Data Statistics}
We computed the average number of words in the text context for all samples within each task type using the \texttt{spaCy} library.\footnote{We use the \texttt{spaCy} library in Python, available at \url{https://pypi.org/project/spacy/}.}
As shown in Table~\ref{tab:dataset_statistics}, while there are some variations in text length across tasks, the differences are relatively minor. 
This indicates that text length is unlikely to be a confounding factor in evaluating modality preference across different task types.

\subsection{Illustrative Samples from the MC\textsuperscript{2} Benchmark}
\label{supp:case}
\begin{figure}[htbp]
  \centering
  \begin{subfigure}[b]{0.48\textwidth}
    \centering
    \textbf{\textless image\textgreater{} is a placeholder for below image}\\[1ex]
    \includegraphics[width=\textwidth]{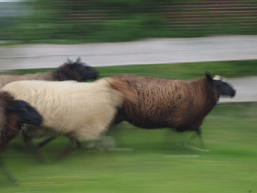}
    \caption*{
    \textbf{User:} \textless image\textgreater \textbf{Conflict Text Context:}  \textit{\textcolor[RGB]{202,12,22}{Three sheep are peacefully eating grass}}, surrounded by lush greenery. Their heads are lowered as they nibble on the fresh blades, completely undisturbed. \textbf{Question:} What are the cows in the back doing? \\
    \textbf{Assistant:} \textless output\textgreater \\
      \textbf{vision-based Answer:} \textit{running}\\
      \textbf{Text-based Answer:} \textit{\textcolor[RGB]{202,12,22}{eating}}
    }
  \end{subfigure}
  \hfill
  \begin{subfigure}[b]{0.48\textwidth}
    \centering
    \textbf{\textless image\textgreater{} is a placeholder for below image}\\[1ex]
    \includegraphics[width=\textwidth]{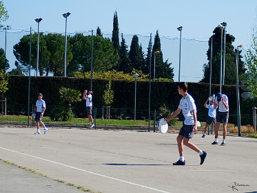}
    \caption*{
    \textbf{User:} \textless image\textgreater \textbf{Conflict Text Context:} In the photo, there are three boys playing Frisbee, and one more boy is partially visible in the corner, bending down to tie his shoelaces, \textit{\textcolor[RGB]{202,12,22}{making a total of four people}}. \textbf{Question:} How many people are in the photo?\\
    \textbf{Assistant:} \textless output\textgreater \\
   \textbf{ vision-based Answer:} \textit{five}\\
    \textbf{Text-based Answer:} \textit{\textcolor[RGB]{202,12,22}{four}}
    }
  \end{subfigure}
\caption{Illustration of using modality context conflict pairs to investigate modality preference in activity recognition (Left) and counting tasks (Right). The highlighted areas indicate the points of conflict between visual and textual contexts.}
\label{fig:case1}
\end{figure}

\begin{figure}[htbp]
  \centering
  \begin{subfigure}[b]{0.48\textwidth}
    \centering
    \textbf{\textless image\textgreater{} is a placeholder for below image}\\[1ex]
    \includegraphics[width=\textwidth]{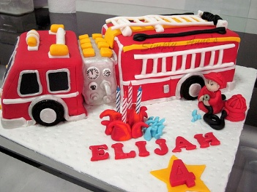}
    \caption*{
    \textbf{User:} \textless image\textgreater \textbf{Conflict Text Context:} \textit{\textcolor[RGB]{202,12,22}{The birthday cake was designed to look like a sleek police car}}, complete with edible flashing lights and a fondant badge on the side. \textbf{Question:} What is the cake in the shape of? \\
    \textbf{Assistant:} \textless output\textgreater \\
      \textbf{vision-based Answer:} \textit{fire truck }\\
      \textbf{Text-based Answer:} \textit{\textcolor[RGB]{202,12,22}{police car}}
    }
  \end{subfigure}
  \hfill
  \begin{subfigure}[b]{0.48\textwidth}
    \centering
    \textbf{\textless image\textgreater{} is a placeholder for below image}\\[1ex]
    \includegraphics[width=\textwidth]{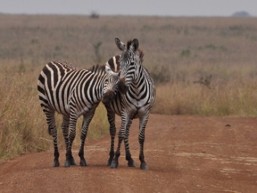}
    \caption*{
    \textbf{User:} \textless image\textgreater \textbf{Conflict Text Context:} \textit{\textcolor[RGB]{202,12,22}{Two wildebeests are standing in a dry}}, grass-less savanna, their dark coats contrasting with the dusty ground. The area is sparse, with only a few scattered shrubs visible in the background.
     \textbf{Question:} What animal is shown?\\
    \textbf{Assistant:} \textless output\textgreater \\
   \textbf{ vision-based Answer:} \textit{zebras}\\
    \textbf{Text-based Answer:} \textit{\textcolor[RGB]{202,12,22}{wildebeests}}
    }
  \end{subfigure}
\caption{Illustration of using modality context conflict pairs to investigate modality preference in attribute recognition (Left) and object recognition tasks (Right). The highlighted areas indicate the points of conflict between visual and textual contexts.}
\label{fig:case2}
\end{figure}

\begin{figure}[htbp]
  \centering
  \begin{subfigure}[b]{0.48\textwidth}
    \centering
    \textbf{\textless image\textgreater{} is a placeholder for below image}\\[1ex]
    \includegraphics[width=\textwidth]{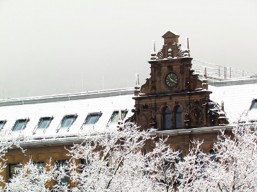}
    \caption*{
    \textbf{User:} \textless image\textgreater \textbf{Conflict Text Context:}  A large brown clock tower mounted in the face of\textit{\textcolor[RGB]{202,12,22}{a building overlooks a vibrant park filled with lush green trees}}. The contrast between the brown tower and the surrounding greenery creates a picturesque scene. \textbf{Question:} What color are the trees? \\
    \textbf{Assistant:} \textless output\textgreater \\
      \textbf{vision-based Answer:} \textit{white}\\
      \textbf{Text-based Answer:} \textit{\textcolor[RGB]{202,12,22}{green}}
    }
  \end{subfigure}
  \hfill
  \begin{subfigure}[b]{0.48\textwidth}
    \centering
    \textbf{\textless image\textgreater{} is a placeholder for below image}\\[1ex]
    \includegraphics[width=\textwidth]{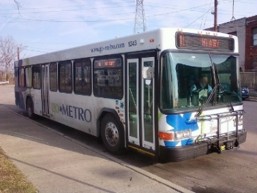}
    \caption*{
    \textbf{User:} \textless image\textgreater \textbf{Conflict Text Context:} A white bus with a large rack on the front is parked by the beach, designed to \textit{\textcolor[RGB]{202,12,22}{carry equipment for surfing}} .The rack is sturdy and spacious, perfect for securing bulky items. \textbf{Question:} What can you hang on the rack on the front of the bus?\\
    \textbf{Assistant:} \textless output\textgreater \\
   \textbf{ vision-based Answer:} \textit{bikes}\\
    \textbf{Text-based Answer:} \textit{\textcolor[RGB]{202,12,22}{surfboards}}
    }
  \end{subfigure}
\caption{Illustration of using modality context conflict pairs to investigate modality preference in color recognition (Left) and positional reasoning (Right) tasks. The highlighted areas indicate the points of conflict between visual and textual contexts.}
\label{fig:case3}
\end{figure}

\begin{figure}[htbp]
  \centering
  \begin{subfigure}[b]{0.48\textwidth}
    \centering
    \textbf{\textless image\textgreater{} is a placeholder for below image}\\[1ex]
    \includegraphics[width=\textwidth]{figure/case/sentiment.jpg}
    \caption*{
    \textbf{User:} \textless image\textgreater \textbf{Conflict Text Context:}  A girl sitting at a counter with a piece of pizza, staring blankly at the wall while the pizza grows cold in front of her. The room is quiet, and \textit{\textcolor[RGB]{202,12,22}{she seems uninterested  in her surroundings}}. \textbf{Question:} What is the girl on the right feeling in the image? \\
    \textbf{Assistant:} \textless output\textgreater \\
      \textbf{vision-based Answer:} \textit{happy}\\
      \textbf{Text-based Answer:} \textit{\textcolor[RGB]{202,12,22}{bored}}
    }
  \end{subfigure}
  \hfill
  \begin{subfigure}[b]{0.48\textwidth}
    \centering
    \textbf{\textless image\textgreater{} is a placeholder for below image}\\[1ex]
    \includegraphics[width=\textwidth]{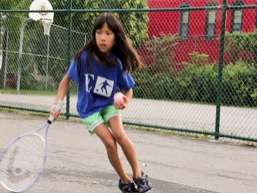}
    \caption*{
    \textbf{User:} \textless image\textgreater \textbf{Conflict Text Context:} The young girl is running swiftly across the field, \textit{\textcolor[RGB]{202,12,22}{ dribbling a soccer ball with precision}} as she maneuvers past imaginary opponents. Her focus is on scoring a goal, and she practices her footwork with determination. \textbf{Question:} What sport is depicted in the picture?\\
    \textbf{Assistant:} \textless output\textgreater \\
   \textbf{ vision-based Answer:} \textit{tennis}\\
    \textbf{Text-based Answer:} \textit{\textcolor[RGB]{202,12,22}{soccer}}
    }
  \end{subfigure}
\caption{Illustration of using modality context conflict pairs to investigate modality preference in sentiment understanding and object recognition
 tasks. The highlighted areas indicate the points of conflict between visual and textual contexts.}
 \label{fig:case4}
\end{figure}

To provide an intuitive understanding of the  MC\textsuperscript{2} benchmark and the nature of modality conflict, we present a few representative samples covering different task types as shown in Figure~\ref{fig:case1}, Figure~\ref{fig:case2}, Figure~\ref{fig:case3} and Figure~\ref{fig:case4}.

\section{Model Evaluatiion}
\label{supp:evaluation}

\subsection{Evaluatiion Detail for Modality Preference}
We assess open-source multi-modal large language models (MLLMs) with different parameter sizes, including LLaVA1.5-7B/13B~\citep{llava1.5}, LLama3.2-11B-Vision-Instruct~\citep{llama3.2}, OneVision-7B/72B~\citep{onevision}, CogVLM2-19B~\citep{cogvlm2}, mPLUG-Owl3-24-07~\citep{mplug}, Qwen2VL-7B~\citep{qwen2vl}, GLM-4V-9B~\citep{du2022glm}, SPHINX-V2-1K~\citep{lin2023sphinx}, InternVL3-9B/14B/38B/78B~\citep{chen2024internvl}, LLaVA-next-7B/13B/34B~\citep{llava1.5} and Qwen2.5VL-7B/32B/72B~\citep{qwen2.5vl}. 
All the open-source models are evaluated using NVIDIA A100 or A800 GPUs.
We also evaluate the proprietary model, GPT-4o-mini~\citep{gpt4o} via the official API.

\paragraph{Details of single-modality context evaluation}
\label{supp:single_modality}
Before evaluating modality preference, we first assess the ability of MLLMs to answer questions accurately given a single-modality context in the MC\textsuperscript{2} dataset. Specifically, we evaluate the models' accuracy in answering based on text context and based on vision context (based on the image). As shown in Table~\ref{tab:base_text} and Table~\ref{tab:base_vision}, all models achieve over 95\% accuracy when provided with either textual or visual context.
This indicates that question understanding and the understanding of single-modality context do not affect the modality preference evaluation. Therefore, we have excluded this confounding factor from the analysis.

\paragraph{Details of results for modality preference evaluation}
We provide the results of modality preference for several models in the left panel of Figure~\ref{fig:single} in the main text. 
More detailed modality preference evaluation results are presented in Table~\ref{tab:single_full}.

\subsection{Can the Vision Ratio provide guidance for downstream task performance?}
\label{supp:vision_ratio_dis_post}
We evluate the performance of Qwen2.5VL-7B, Qwen2.5VL-72B, InternVL3-9B, InternVL3-14B, InternVL3-38B, InternVL3-78B, LLaVA-OneVison-7B, LLaVA-OneVison-72B, LLava-next-7B, LLava-next-13B on 7 general multi-modal understanding benchmarks including MMMU~\citep{mmmu}, MME~\citep{mme}, MMBench~\citep{mmbench}, RealwordQA~\citep{infovqa}, MMStar~\citep{chartqa}, InfoVQA~\citep{infovqa} and ChartQA~\citep{chartqa}.
We compute the average score on all datasets, where MME score is normalized between 0-1, as shown in Table~\ref{tab:all}.

\begin{table*}[ht]
\centering
\small
\setlength{\tabcolsep}{6pt} 
\renewcommand{\arraystretch}{1.2} 
\resizebox{0.95\linewidth}{!}{
\begin{tabular}{lccccccccc c}
\toprule
\textbf{Model} & \textbf{MMMU} & \textbf{MME} & \textbf{MMBench} & \textbf{RealworldQA} & \textbf{MMStar} & \textbf{HallBench} & \textbf{InfoVQA} & \textbf{ChartQA} & \textbf{Avg} & \textbf{Vision Ratio} \\
\midrule
Qwen2.5VL-7B        & 58.6 & 83.8 & 83.5 & 68.5 & 63.9 & 52.9 & 82.6 & 87.3 & 75.5 & 59.6 \\
Qwen2.5VL-72B       & 70.2 & 87.4 & 88.6 & 75.7 & 70.8 & 55.2 & 87.3 & 89.5 & 81.4 & 78.6 \\
InternVL3-9B        & 57.7 & 84.7 & 83.4 & 70.5 & 66.3 & 51.2 & 79.6 & 86.2 & 75.5 & 41.2 \\
InternVL3-14B       & 67.1 & 88.5 & 85.6 & 70.7 & 68.8 & 55.1 & 83.6 & 87.3 & 78.8 & 55.0 \\
InternVL3-38B       & 70.1 & 90.1 & 87.6 & 75.6 & 71.5 & 57.1 & 85.0 & 89.2 & 81.3 & 62.4 \\
InternVL3-78B       & 72.2 & 91.1 & 89.0 & 78.0 & 72.5 & 59.1 & 86.5 & 89.7 & 82.7 & 81.5 \\
OneVision-7B  & 47.9 & 71.2 & 83.2 & 66.3 & 61.7 & 31.6 & 68.8 & 80.0 & 68.4 & 26.3 \\
LLaVA-OneVision-72B & 55.7 & 80.8 & 85.8 & 71.9 & 65.8 & 49.0 & 74.9 & 83.7 & 74.1 & 30.1 \\
Qwen2VL-7B          & 54.1 & 83.1 & 83.0 & 70.1 & 60.7 & 50.6 & 76.5 & 83.0 & 72.9 & 16.3 \\
LLaVA-Next-7B       & 37.6 & 63.2 & 69.2 & 57.8 & 37.6 & 27.6 & 31.6 & 51.9 & 49.8 & 8.5  \\
LLaVA-Next-13B      & 37.3 & 62.3 & 70.0 & 57.6 & 40.4 & 31.8 & 34.9 & 59.0 & 51.6 & 9.7  \\
LLaVA-1.5-7B        & 35.7 & 64.6 & 69.2 & 54.8 & 33.1 & 27.6 & 22.4 & 17.8 & 42.5 & 13.4 \\
LLaVA-1.5-13B       & 37.0 & 63.6 & 66.5 & 55.3 & 34.3 & 24.5 & 24.9 & 18.5 & 42.9 & 15.0 \\
\bottomrule
\end{tabular}
}
\caption{Performance comparison across benchmarks for different models measured by accuracy (\%) and Vision Ratiio score (\%).}
\label{tab:all}
\end{table*}

\subsection{The Details for Controlling Modality Preference}
\label{supp:control}

\paragraph{More results for controlling modality preference through instruction design.}
In the left panel of Figure~\ref{fig:pca}, we provide the Vision Ratio results for OneVision-7B, Qwen2.5VL-7B, Qwen2VL-7B and InternVL3-8B.
We also present more results on controlling modality preference through instruction design for preference towards the vision modality and the text modality in Table~\ref{tab:icc_img} and Table~\ref{tab:icc_text}.
For each setting, we report the results measured by $S_{\text{vision}}$, vision-based accuracy and $S_{\text{text}}$, text-based accuracy.

\paragraph{The details of PCA Analysis}
In Section~\ref{sec:interventions}, we use the PCA analysis regarding the Modality Preference Direction in Representation Space. 
Here, we provide a more detailed description of the setup.
We extract the model's hidden representations from the last token of the input across different layers. Then, we apply the PCA method to reduce the dimensionality to two dimensions for visualization.
The following settings were visualized:
\begin{enumerate}
    \item The model states under the original modality context input in conflicting scenarios.
    \item The model states when there is image noise or textual syntax errors.
    \item The model states when specific instructions biased towards image or text are added.
\end{enumerate}

To improve PCA dimensionality reduction efficiency, we selected 500 samples for each setting. 
Additionally, we calculated the center position after dimensionality reduction for each setting. 
The center (or centroid) of the samples is computed by taking the mean of the reduced-dimensional points across all the samples.

\section{Method Applying}
\label{supp:applying}
\subsection{Details for Pattern of hidden states}
\label{sec:pattern}
\begin{figure*}[t]
  \subfloat[OneVision-7B]{\includegraphics[height=0.24\textwidth]{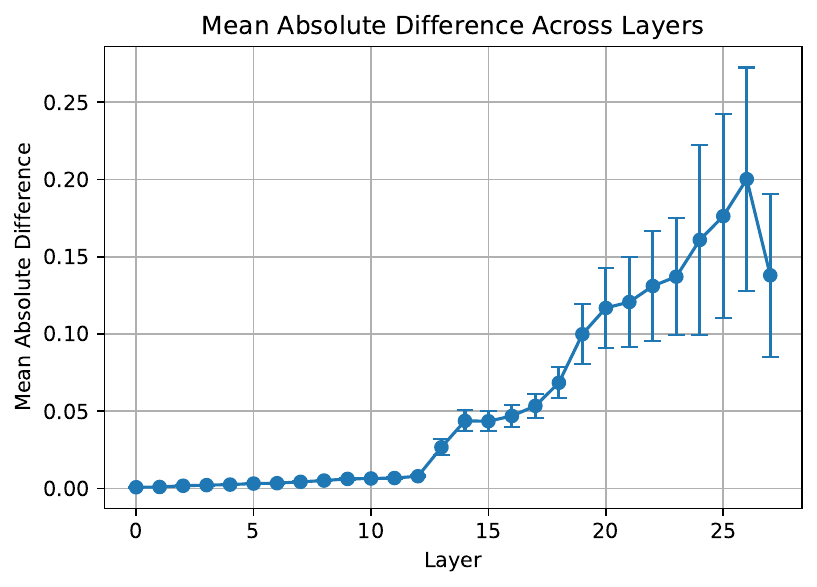}} \hfill
  \subfloat[Qwen2VL-7B]{\includegraphics[height=0.24\textwidth]{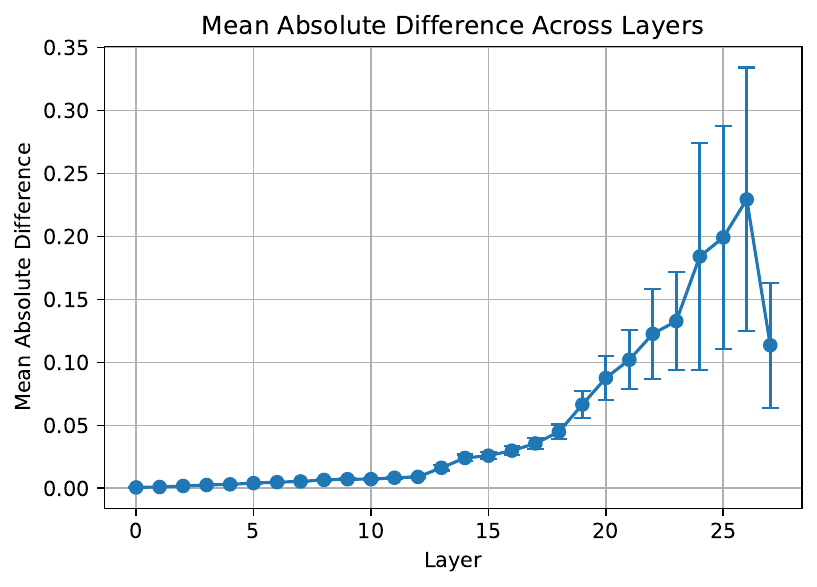}} \hfill
  \subfloat[InternVL3-8B]{\includegraphics[height=0.24\textwidth]{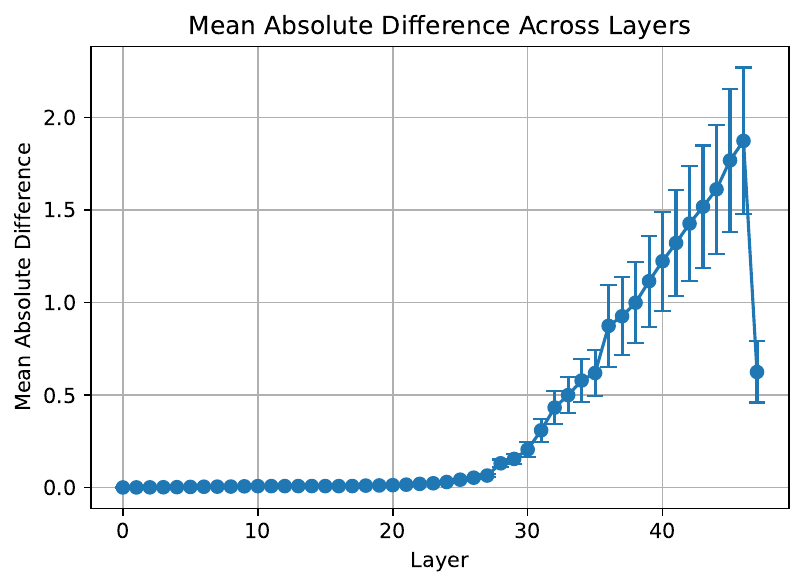}} 
  \caption{
    Layer-wise absolute difference and standard deviation of hidden states between image-guided and text-guided instruction for OneVision-7B, Qwen2VL-7B and InternVL3-8B models from left to right.
  }
  \label{fig:other_layer_wise}
\end{figure*}

In the main text, we visualize the layer-wise absolute difference and standard deviation of the hidden states for Qwen2.5VL-7B. 
As shown in Figure~\ref{fig:other_layer_wise}, we present the visualization of hidden states for OneVision-7B, Qwen2VL-7B, and InternVL3-8B.
For each model, we selected layers with large absolute differences and small standard deviations. 
This means we identified the layers that showed stable and significant differences between instructions with modality preference towards vision context and text context, which are then used to steer and adjust the model's modality preference.

\subsection{Evaluation of visual understanding and multi-modal multi-modal machine translation}
PhD~\citep{phd} is a visual understanding benchmark and includes two subsets—PhD-ica, which contains irrelevant textual context, and PhD-icc, which introduces misleading or incorrect textual information—both of which increase the risk of hallucination.
For testing convenience, we randomly selected 1,000 samples from the original Phd-cc and Phd-ica datasets for evaluation.
By steering the model’s modality preference toward the vision modality, we strengthen its visual understanding ability and mitigate vision hallucinations in MLLMs. \

Ambigcaps~\citep{ambigcaps} benchmark explores the role of datasets in stimulating the leverage of the visual modality and proposes methods to highlight the importance of visual signals in the datasets. 
We evaluate the multi-modal machine translation (MMT) task on this dataset using the Qwen2.5VL-7B model.
multi-modal contexts in MMT are both complementary and contradictory: the visual information provides helpful context for translation, but the potential for conflicting, non-visual signals can interfere with grounding the source language. Consequently, the proposed method is designed to steer the modality preference toward the text modality to ensure robustness against these visual-textual conflicts.

Conversely, when guided toward the text modality, the model places greater emphasis on the source language, leading to more accurate grounding in multi-modal machine translation. This adjustment prevents the model from over-relying on visual content and from introducing spurious objects or extraneous details into the translation output.

\section{More Experiment Analysis}
\label{supp:more_exp}

\subsection{The Sensitivity Analysis for the Evaluation of Modality Preference}
\label{supp:number}
To verify whether the current \textbf{2k-sample scale} of $MC^2$ is sufficient to ensure the stability of both the preference evaluation and steering results, we conduct a sensitivity analysis by randomly selecting a specified quantity of samples from each category for assessment.
As shown in Table~\ref{tab:sensitivity_benchmark}, we calculate the Vision Ratio for OneVision-7B and Qwen2.5VL-7B. 
Results demonstrate that as the sample size increases per category, the Vision Ratio begins to stabilize around 150 samples. 
These experiments suggest that the current dataset size for each task is sufficient to ensure the stability of evaluation of modality preference.

In the future, we would like to expand MC2 with a wider variety of tasks (e.g., texture recognition) and modalities (e.g., Audio), further enhancing its comprehensiveness and generalizability.

\begin{table}[h]
    \centering
    \caption{The sensitivity analysis for the evaluation of modality preference. We  randomly select a specified quantity of samples from each category for assessment, measured by Vision Ratio.}
    \label{tab:sensitivity_benchmark}
    \begingroup %
    \begin{tabular}{l cccccccccc}
        \toprule
        \textbf{Model} & \textbf{25} & \textbf{50} & \textbf{75} & \textbf{100} & \textbf{125} & \textbf{150} & \textbf{175} & \textbf{200} & \textbf{225} & \textbf{250} \\
        \midrule
        OneVision-7B & 29.2 & 28.0 & 29.1 & \textbf{29.6} & 28.1 & 26.4 & 26.6 & 26.7 & 26.7 & 26.3 \\
        Qwen2.5VL-7B & 56.7 & 59.4 & 60.2 & \textbf{59.9} & 60.4 & 59.8 & 60.2 & 59.7 & 59.8 & 59.6 \\
        \bottomrule
    \end{tabular}
    \endgroup 
\end{table}

\subsection{More Results for Downstream Tasks}
\label{supp:more_results_}
\subsubsection{More results for Phd dataset}
We evaluate the performance of Phd dataset using the proposed method for LLaVA-1.6-7B and the results are in Table~\ref{tab:phd_llava16}.
We observe that LLaVA-1.6-7B, due to its \textbf{severe text preference}, only achieves an average ACC score of $0.7$ on the PHD-icc subset. 
Applying our method significantly boosts the model's performance across the two subsets, thereby demonstrating the cross-model generalization of our approach.

\begin{table}[t]
    \centering
    \caption{Performance of the proposed method on the visual understanding benchmark, Phd using LLaVA-1.6-7B.} 
    \label{tab:phd_llava16}
    \begingroup
    \scalebox{1.0}{ 
        \begin{tabular}{ll cccccc}
            \toprule
            \textbf{Phd} & \textbf{Method} & \textbf{Attribute} & \textbf{Sentiment} & \textbf{Positional} & \textbf{Counting} & \textbf{Object} & \textbf{Avg} \\
            \midrule
            \multirow{3}{*}{\textbf{Phd-icc}}  
             & LLaVA-1.6-7B & 0.5 & 0.0 & 0.0 & 1.5 & 1.5 & 0.7 \\
             & InstDesign & 2.0 & 0.5 & 0.5 & 1.5 & 9.5 & 2.8 \\
             & \textbf{Ours} & \textbf{3.5} & \textbf{1.5} & \textbf{1.0} & \textbf{3.0} & \textbf{15.0} & \textbf{4.8} \\
            \midrule %
            \multirow{3}{*}{\textbf{Phd-iac}}  
             & LLaVA-1.6-7B & 5.5 & 8.0 & 4.0 & 10.5 & 29.0 & 11.4 \\
             & InstDesign & 7.5 & 12.5 & 14.5 & 15.5 & 44.5 & 18.9 \\
             & \textbf{Ours} & \textbf{11.5} & \textbf{20.0} & \textbf{20.5} & \textbf{22.0} & \textbf{51.0} & \textbf{25.0} \\
            \bottomrule
        \end{tabular}
    }
    \endgroup
\end{table}

\subsection{Ablation Study}
\label{supp:ablation}
In this section, we conduct the detailed ablation study to analyze the proposed method.
\subsubsection{The number of probing samples}
We compute the preference direction in Equation~\ref{eq:probing} using varied sample sizes but test the steering performance on the complete $MC^2$ dataset. 
We report the $S_{\text{Vision}}$ and $S_{\text{Text}}$ results for  OneVision-7B and Qwen2.5VL-7B in Table~\ref{tab:probing_number}. 
We observe that steering performance remains stable even when the steering vector is derived from a limited number of samples.
\begin{table}[h]
    \centering
    \caption{The ablation study for the varied sample number of computing the preference direction.}
    \label{tab:probing_number}
    \begingroup %
    \begin{tabular}{l cccccccccc}
        \toprule
        \textbf{Model} & \textbf{25} & \textbf{50} & \textbf{75} & \textbf{100} & \textbf{125} & \textbf{150} & \textbf{175} & \textbf{200} & \textbf{225} & \textbf{250} \\
        \midrule
        OneVision-7B & 56.0 & 55.9 & 56.3 & \textbf{57.1} & 57.0 & 57.3 & 57.3 & 57.4 & 57.2 & 57.1 \\
        Qwen2.5VL-7B & 62.2 & 62.2 & 61.7 & \textbf{62.6} & 64.0 & 62.8 & 62.7 & 63.1 & 62.8 & 63.6 \\
        \bottomrule
    \end{tabular}
    \endgroup %
\end{table}
\subsubsection{The diversity of probing samples}
To study the impact of data diversity for probing task, we experiment by using \textbf{only a single task for probing} and applying the resulting vector to steer \textbf{all other tasks}. We report the $S_{\text{Vision}}$ and $S_{\text{Text}}$ for OneVision-7B and Qwen2.5VL-7B for entire dataset in Table~\ref{tab:probing_diveristy}.
We observe that OneVision-7B achieves competitive performance compared to our initial implementation for using nearly each probing task, with Qwen2.5VL-7B showing similar success on over half the tasks. Further analysis finds that the most effective single-probing tasks are those where the initial modality preference change was \textbf{more pronounced}.

\begin{table}[h]
    \centering
    \caption{The ablation study for the varied diversities of computing the preference direction.}
    \label{tab:probing_diveristy}
    \begingroup %
        \scalebox{0.75}{
    \begin{tabular}{l ccccccccc}
        \toprule
        \textbf{Model} & \textbf{Sport} & \textbf{Attribute} & \textbf{Sentiment} & \textbf{Positional} & \textbf{Counting} & \textbf{Color} & \textbf{Activity} & \textbf{Object} & \textbf{Ours} \\
        \midrule
        OneVision-7B & 58.6 & 55.8 & 59.3 & 57.8 & 59.0 & 53.4 & 58.5 & 58.6 & 57.1 \\
        Qwen2.5VL-7B & 64.1 & 59.5 & 70.0 & 51.9 & 47.4 & 65.5 & 63.0 & 60.6 & 63.6 \\
        \bottomrule
    \end{tabular}
    }
    \endgroup 
\end{table}

\subsubsection{Different steering intensities}
To investigate the performance with varied steering intensities, we introduce a scaling coefficient $\lambda$ to the steering weight $w$ in Equation~\ref{eq:steering} to change the steering intensity and conduct a test on $MC^2$ for both OneVision-7B and Qwen2.5-VL-7B, reporting the $S_{\text{Vision}}$ or $S_{\text{Text}}$ scores in Table~\ref{tab:steer_intensity}.
We observe that performance drops for both models with decreased steering intensity, indicating \textbf{insufficient steering}. 
As intensity increases, the performance of OneVision-7B significantly drops, exhibiting clear \textbf{over-steering} at $\lambda=2.0$ which leads to destruction of language capabilities.
Conversely, for Qwen2.5VL-7B, the steering effect continues to enhance up to $\lambda=1.75$, only significantly degrading beyond $\lambda=2.0$, demonstrating a wider \textbf{safe steering margin} in its representation space.  
\begin{table}[h]
    \centering
    \caption{The ablation study for the varied steering intensities.}
    \label{tab:steer_intensity}
    \begingroup %
    \scalebox{0.8}{
    \begin{tabular}{l cccccccccccccc}
        \toprule
        \textbf{Model} & \textbf{0.125} & \textbf{0.25} & \textbf{0.5} & \textbf{0.75} & \textbf{1.0} & \textbf{1.25} & \textbf{1.5} & \textbf{1.75} & \textbf{2.0} & \textbf{2.25} & \textbf{2.5} & \textbf{2.75} & \textbf{3.0} \\
        \midrule
        OneVision-7B & 40.2 & 45.5 & 51.8 & 52.6 & 57.1 & 56.5 & 32.6 & 12.1 & 3.9 & 0.1 & 0.0 & 0.0 & 0.0 \\
        Qwen2.5VL-7B & 40.8 & 44.1 & 49.4 & 53.2 & 63.6 & 72.3 & 76.0 & 70.7 & 45.8 & 15.2 & 16.2 & 12.5 & 7.8 \\
        \bottomrule
    \end{tabular}
    }
    \endgroup 
\end{table}

\begin{figure}[t]
  \centering
  \small
  \textbf{\textless image\textgreater{} represents the visual input displayed below}\\[1ex]
  \includegraphics[width=0.8\textwidth]{figure/case_exp/26/COCO_val2014_000000473541.jpg}
  
  \vspace{1ex}
  \begin{flushleft}
    \setlength{\leftskip}{2em} %
    \textbf{User:} $\langle \text{image} \rangle$ \textbf{Text Context:} The table was adorned with a vibrant bouquet of flowers and a charming ceramic sheep, while the surrounding chairs, crafted from smooth, \textit{\textcolor[RGB]{202,12,22}{polished wood}}, complemented the rustic yet elegant setting. In case of inconsistency, prioritize the text context. \\
    \textbf{Question:} What \textbf{are} the chairs made of? A. wicker \quad B. wood \\
    \textbf{Assistant:} $\langle \text{output} \rangle$  \\[0.5ex]
    \textbf{Vision-based answer:} \textit{wicker}\\
    \textbf{Text-based answer:} \textit{\textcolor[RGB]{202,12,22}{wood}} \\
    \textbf{Baseline answer:} A. wicker \ding{55} \\
    \textbf{Proposed steering answer:} B. wood \checkmark
  \end{flushleft}

  \caption{Example of a cross-modal conflict sample from our diagnostic dataset. The highlighted text indicates the ground-truth modality specified by the instruction, which the proposed steering method successfully follows while the baseline fails.}
  \label{fig:attention_detail}
\end{figure}

\begin{figure*}[t]
  \subfloat{\includegraphics[height=0.24\textwidth]{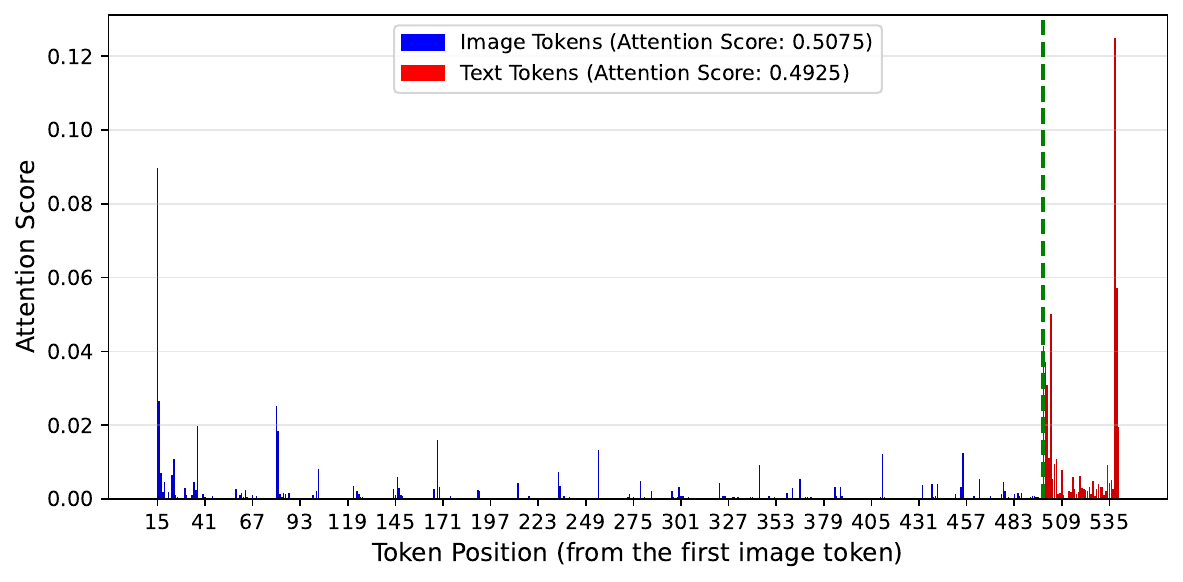}} \hfill
  \subfloat{\includegraphics[height=0.24\textwidth]
  {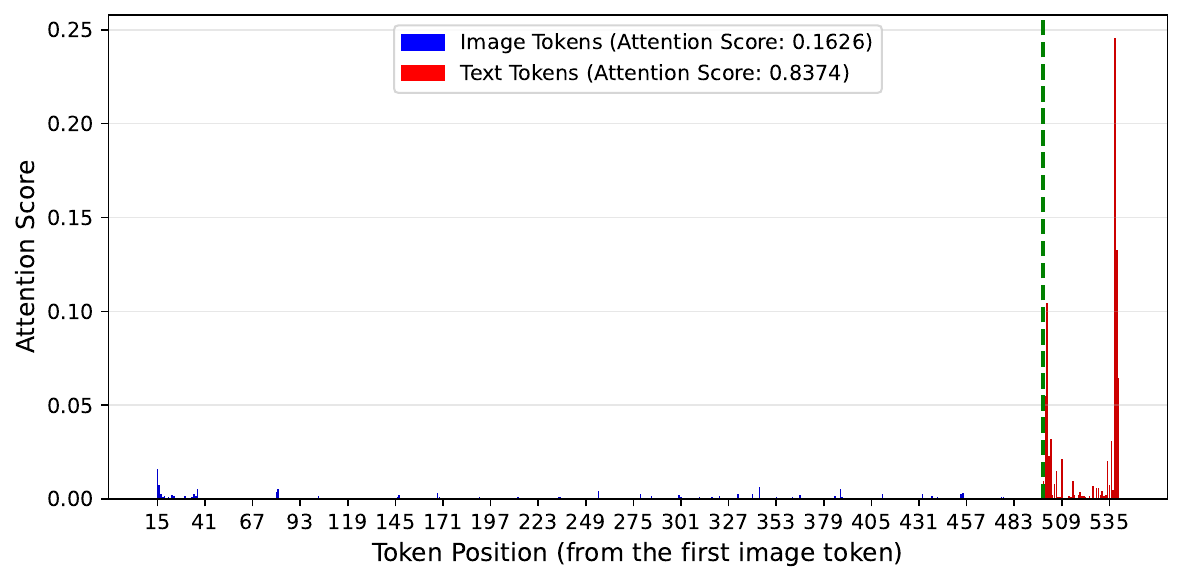}} \hfill
  \caption{The attention scores of the generated token toward the vision and text contexts using InstDesign (Left) and
  the proposed method (Right) in the 23rd layer.
}
  \label{fig:attention_steer23}
\end{figure*}
\begin{figure*}[t]
  \subfloat{\includegraphics[height=0.24\textwidth]{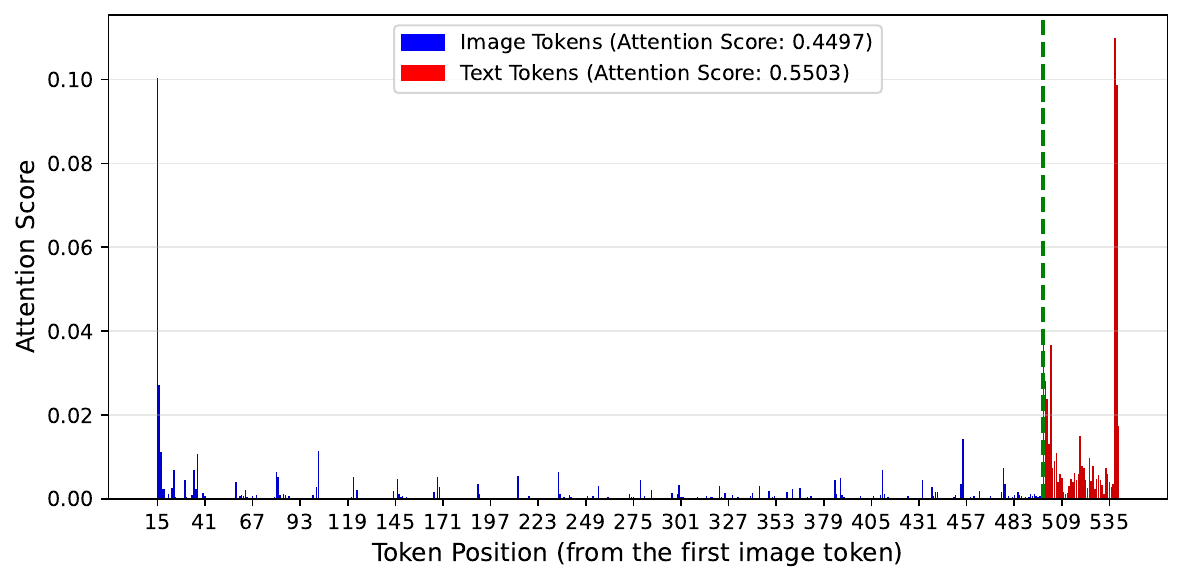}} \hfill
  \subfloat{\includegraphics[height=0.24\textwidth]
  {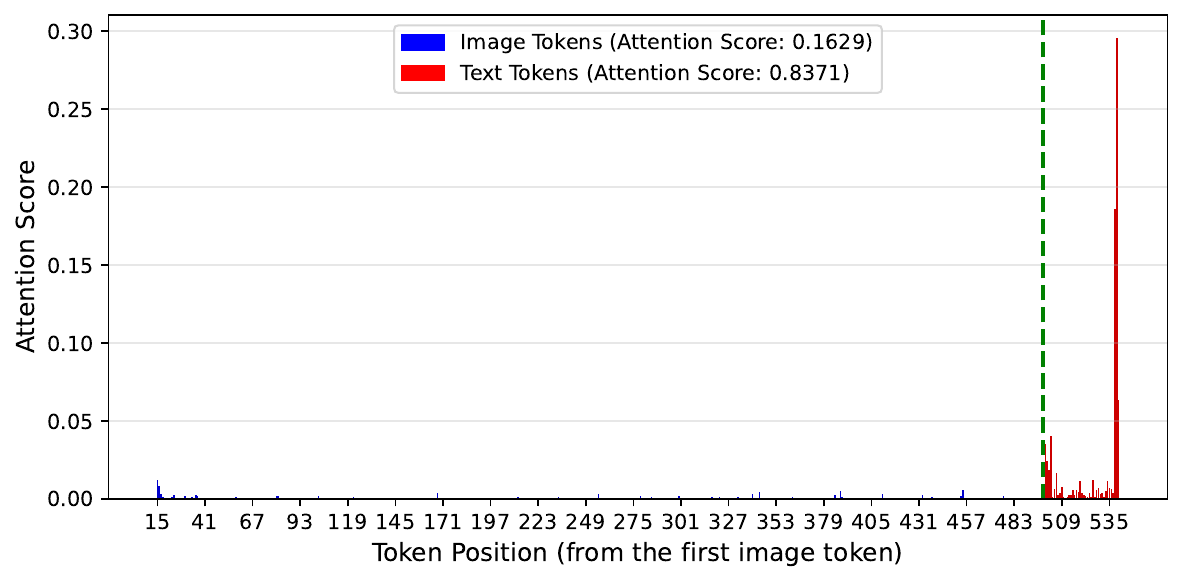}} \hfill
  \caption{The attention scores of the generated token toward the vision and text contexts using InstDesign (Left) and
  the proposed method (Right) in the 24th layer.
}
  \label{fig:attention_steer24}
\end{figure*}
\begin{figure*}[h]
  \subfloat{\includegraphics[height=0.24\textwidth]{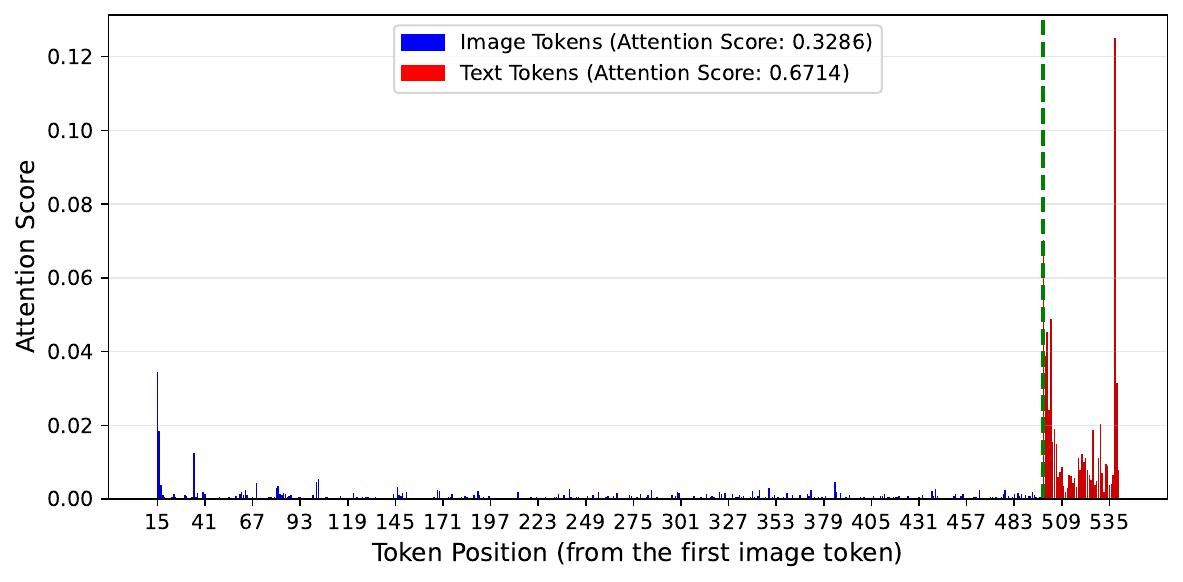}} \hfill
  \subfloat{\includegraphics[height=0.24\textwidth]
  {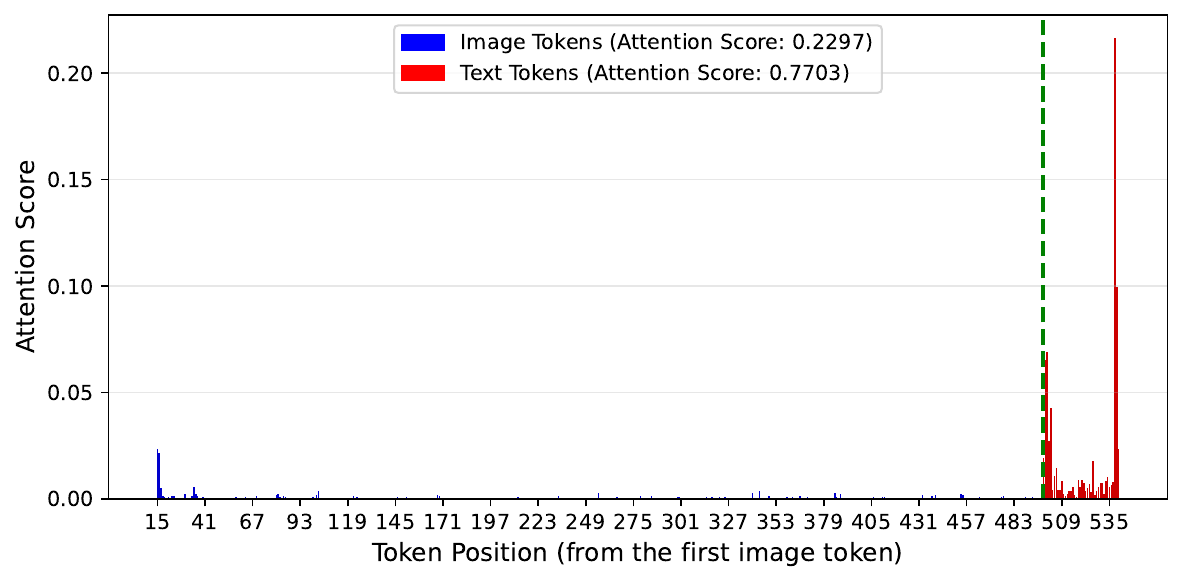}} \hfill
  \caption{The attention scores of the generated token toward the vision and text contexts using InstDesign (Left) and
  the proposed method (Right) in the 25th layer.
}
  \label{fig:attention_steer25}
\end{figure*}
\begin{figure*}[t]
  \subfloat{\includegraphics[height=0.24\textwidth]{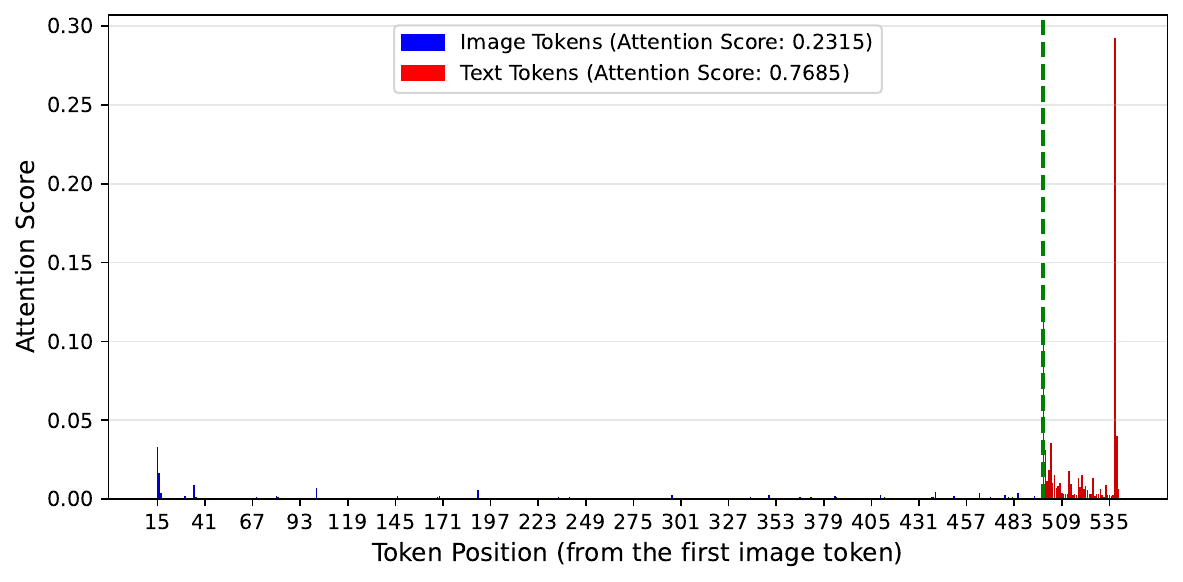}} \hfill
  \subfloat{\includegraphics[height=0.24\textwidth]
  {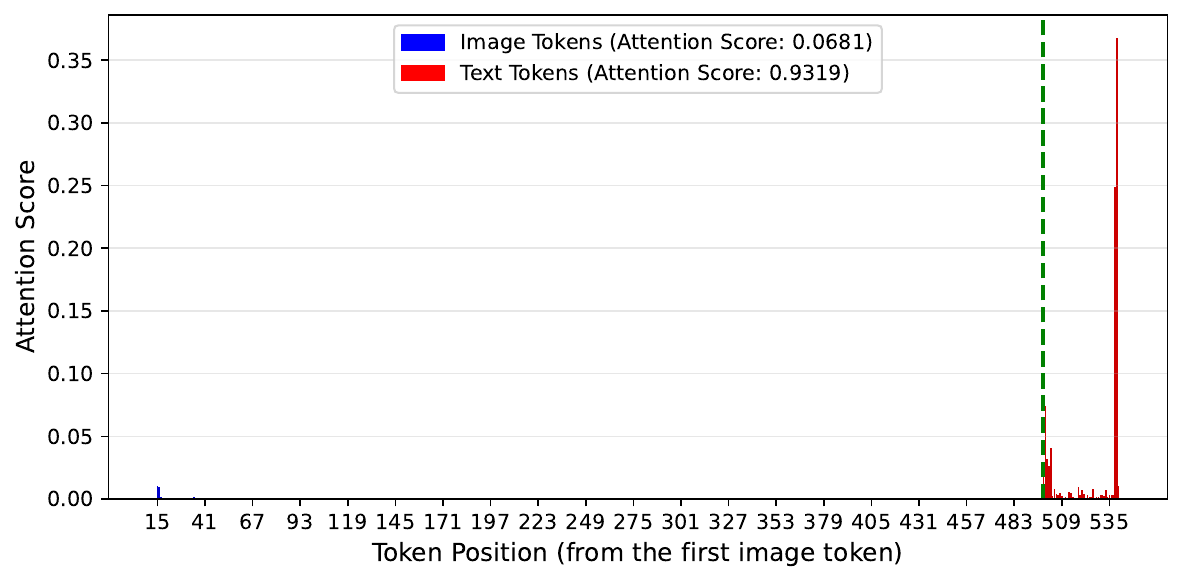}} \hfill
  \caption{The attention scores of the generated token toward the vision and text contexts using InstDesign (Left) and
  the proposed method (Right) in the 26th layer.
}
  \label{fig:attention_steer26}
\end{figure*}
\begin{figure*}[t]
  \subfloat{\includegraphics[height=0.24\textwidth]{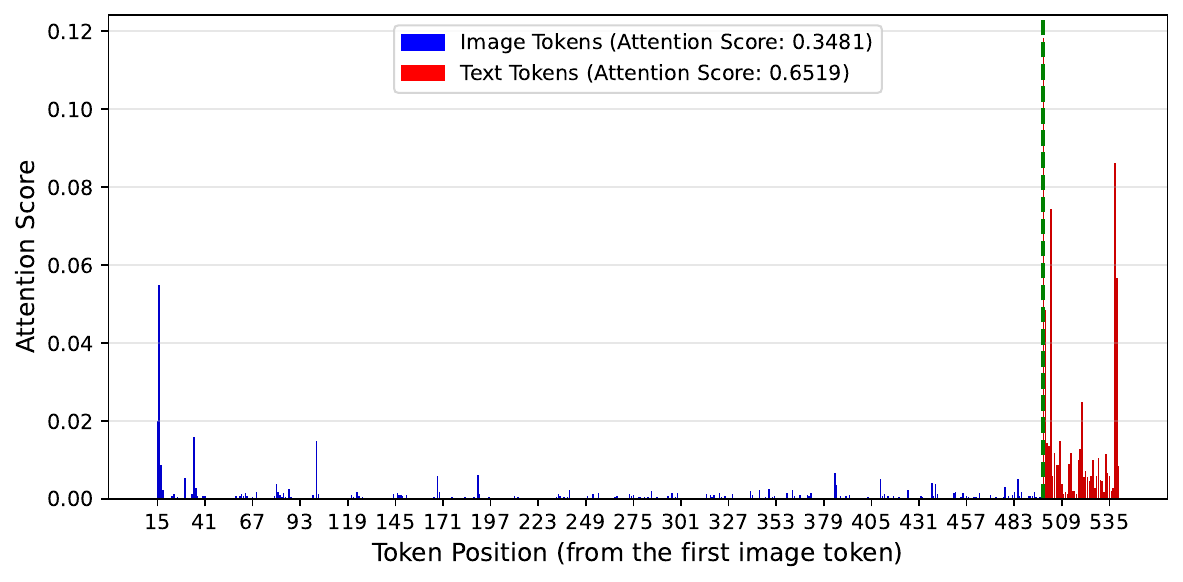}} \hfill
  \subfloat{\includegraphics[height=0.24\textwidth]
  {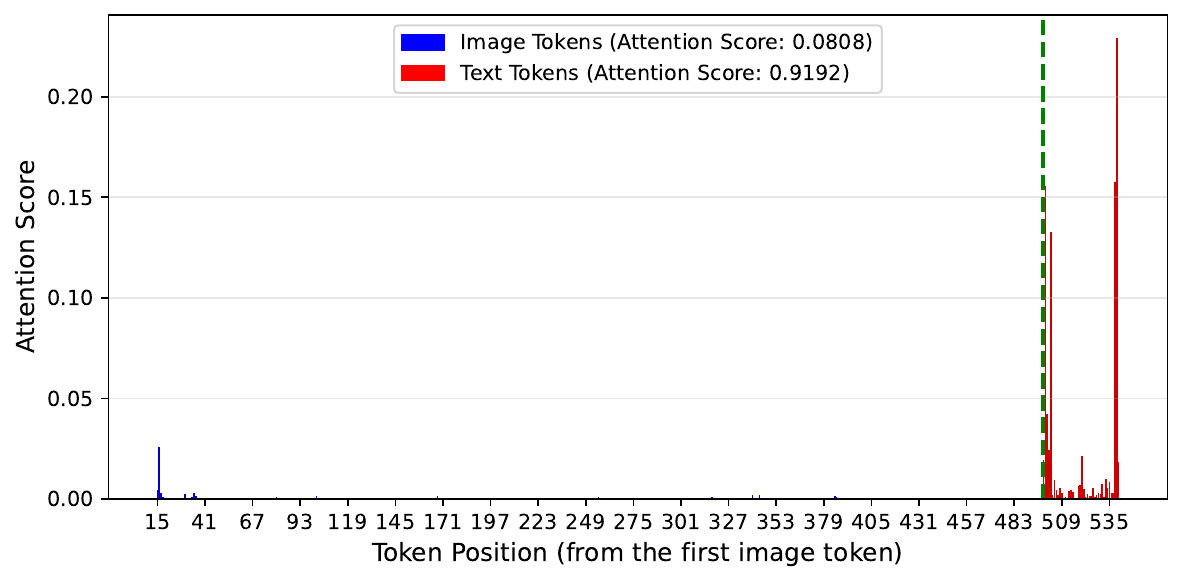}} \hfill
  \caption{The attention scores of the generated token toward the vision and text contexts using InstDesign (Left) and
  the proposed method (Right) in the 27th layer.
}
  \label{fig:attention_steer27}
\end{figure*}
\begin{figure*}[t]
  \subfloat{\includegraphics[height=0.24\textwidth]{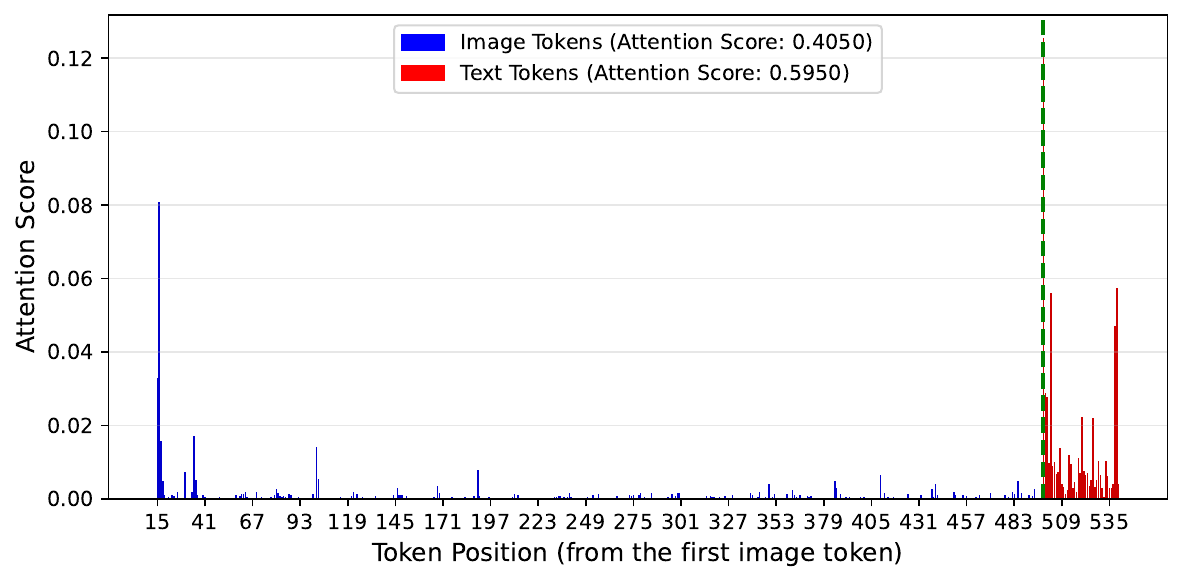}} \hfill
  \subfloat{\includegraphics[height=0.24\textwidth]
  {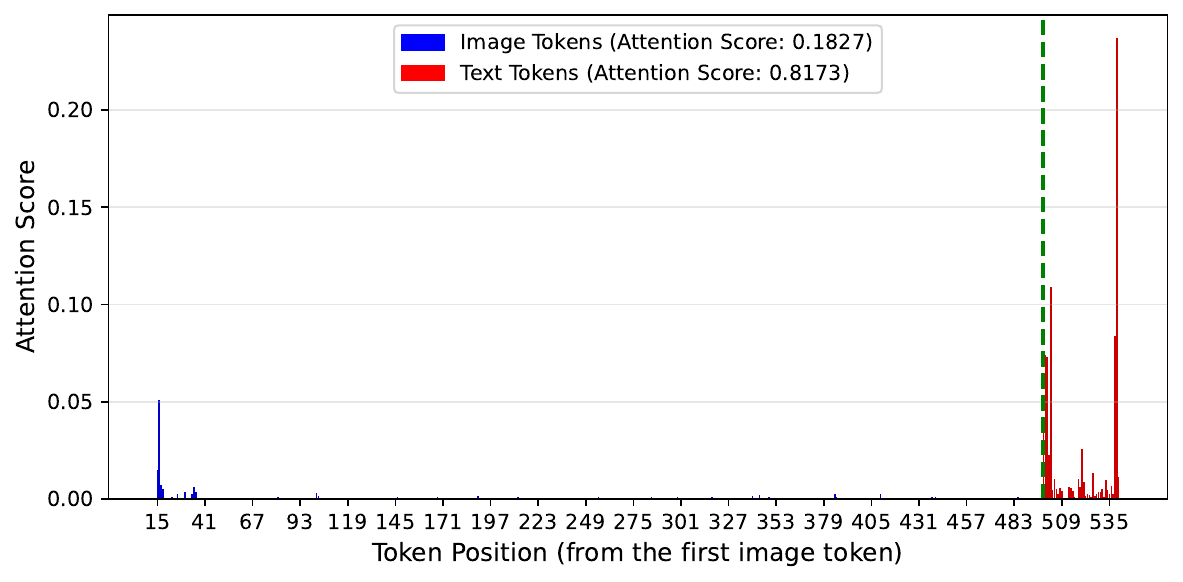}} \hfill
  \caption{The attention scores of the generated token toward the vision and text contexts using InstDesign (Left) and
  the proposed method (Right) in the 28th layer.
}
  \label{fig:attention_steer28}
\end{figure*}
\subsection{Details for the application of the proposed method}
\label{supp:details_analysis}
\subsubsection{Details of in-depth analysis for steering method}
We provide the detailed description of the case for in-depth analysis for steering method in Figure~\ref{fig:attention_steer} in Section~\ref{sec:in_depth}.
Besides, we also provide more attention analysis for the case in the different layers in Figure~\ref{fig:attention_steer23},~\ref{fig:attention_steer24},~\ref{fig:attention_steer25},~\ref{fig:attention_steer26},~\ref{fig:attention_steer27} and~\ref{fig:attention_steer28}.
We observe that across all subsequent layers following steering modality preference towards text at the 21 th layer, our method significantly increases the model's attention weight toward the text modality, surpassing the corresponding vision attention weight. This change clearly demonstrates that the steering mechanism successfully alters the modality preference by enhancing the model's dependency on the text modality.

\subsubsection{Latency and Memory}
The proposed method consists of two phases, probing and steering. 
The probing phase is conducted offline and the resulting steering vector is cached for reuse.
During the actual steering phase, we simply load this cached steering vector, which incurs minimal memory overhead and does not add meaningful computational cost to the inference process. 
We measure the single-sample inference latency (seconds) (without Flash-Attention acceleration and batch inference) for our method compared to the MLLM baseline (MLLM-only) in Table~\ref{tab:time}.
The results show that the steering phase introduces negligible latency compared to the MLLM-only baseline, and the overhead is confined to the initial offline probing stage. Furthermore, all three methods require nearly identical memory requirements. 
\subsubsection{The prerequisites for implementation of the proposed method}
Based on Representation Engineering~\cite{represent1,represent2}, the proposed method requires capturing an \textbf{explicit modality preference direction vector} to realize behavioral adjustment. 
The approach succeeds when such a vector can be reliably extracted, as seen in models like Qwen2.5VL-7B. 
However, the method fails in cases such as LLaVA-1.5-7B, primarily due to its limited ability to follow instructions for preference adjustment, which prevents the capture of a meaningful direction vector.
\begin{table}[h]
    \centering
    \caption{The comparison of inference time between MLLM-only and the proposed method including probing and steering phases.}
    \label{tab:time}
    \begingroup %
    \begin{tabular}{l c c}
        \toprule
        \textbf{Dataset} & \textbf{MC2} & \textbf{Phd-icc} \\
        \midrule
        MLLM-only & 1.99 & 1.84 \\
        Probing (Offline) & 2.21 & 2.12 \\
        Steering & 2.00 & 1.84 \\
        \bottomrule
    \end{tabular}
    \endgroup 
\end{table}
Besides, we observe a localized performance drop in the Attribute subset of the Phd-icc benchmark in Table~\ref{tab:phd_main}. 
We attribute this to the limitation of applying a single global steering vector, which may fail to accommodate \textbf{instance-level granularity}—where fine-grained, sample-specific features are required for optimal alignment. 
Despite these isolated cases, the overall benchmark performance improves, underscoring the effectiveness of our approach, especially considering that it requires no external data or fine-tuning.
\subsection{The Further Experiment without modality preference prior}
Our current approach intentionally select the steering direction based on known task requirements. 
This design has been proven to be pragmatic and effective for many real-world applications where the optimal modality is clear. 
In addition, our method can be readily integrated with a training-free priority detection method to enable dynamic preference selection. To demonstrate this, we conduct the following experiment:

\noindent\textbf{Dataset Construction}: We modify $MC^2$ dataset by degrading the quality of one modality context so that only one modality is reliable, and the ground-truth answer aligns with it. We use QA accuracy to measure model performance on this new dataset.

\noindent\textbf{Task Design:} 1) Each sample requires a specific reliable modality.  2) All samples share the same reliable modality in a task. Each task contains 200 samples. 

\noindent\textbf{Solution:} We apply a causal analysis approach~\cite{parcalabescu2024vision} to identify the reliable modality. 
For each sample, we first measure the change of predicted answer probability when removing either the image or the text context. The larger the drop, the more important that modality is for the given sample. For Task1, we determine the reliable modality for a specific sample by comparing the probability drops. For Task2, by aggregating the reliable modalities across all samples via majority voting, we determine the preferred modality for the specific task.

\noindent\textbf{Results:} For the identification of reliable modality, we achieve an accuracy of $85.3\%$ for all samples in Task1; we reach $100\%$ accuracy for task-level identification in Task2 (thus, performance on Task 2 is equivalent to knowing the steering preference in advance). Next, we evaluate the performance of the proposed method on Task1, measured by QA accuracy in Table~\ref{tab:steering_no_prior}.

The results show that steering with predicted preference yields significant gains over base models and closely matches the performance of the ``preference prior" setting. 
This confirms that our method can be simply adapted to autonomously prioritize modalities based on input quality or task needs. 
\begin{table}[h]
    \centering
    \caption{The performance of the proposed method on the revised $MC^2$ without modality preference prior measured by Accuracy.}
    \label{tab:steering_no_prior}
    \begingroup 
    \begin{tabular}{l c}
        \toprule
        \textbf{Method} & \textbf{Task1} \\
        \midrule
        OneVision-only & 25.4 \\
        +Steering with preference prior & 40.7 \\
        +Steering with predicted preference & 37.5 \\
        \midrule 
        Qwen2.5VL-7B-only & 49.1 \\
        +Steering with preference prior & 62.7 \\
        +Steering with predicted preference & 58.2 \\
        \bottomrule
    \end{tabular}
    \endgroup 
\end{table}

\begin{table}[t]
\centering
\caption{Accuracy of question answering in the MC\textsuperscript{2} dataset when both textual and visual contexts are provided but the instruction does not specify which modality context should be used. Values are reported as vision-based accuracy/text-based accuracy for each model.}
\resizebox{\linewidth}{!}{
\renewcommand\arraystretch{1.2}
\begin{tabular}{lccccccccc}
\toprule
\textbf{Model} & \textbf{Sport} & \textbf{Attribute} & \textbf{Sentiment} & \textbf{Positional} & \textbf{Counting} & \textbf{Color} & \textbf{Activity} & \textbf{Object} & \textbf{Avg} \\
\midrule
\textbf{LLaMAVision-11B} & 31.2/52.4 & 20.4/69.6 & 2.0/93.2 & 21.2/66.8 & 4.0/93.2 & 35.2/47.2 & 10.0/82.4 & 38.8/42.8 & 20.4/68.4 \\
\textbf{LLaVA1.5-7B} & 20.0/59.6 & 8.0/88.0 & 2.0/86.0 & 8.8/75.2 & 1.2/96.0 & 10.8/82.0 & 9.6/78.8 & 35.2/52.0 & 11.9/77.2 \\
\textbf{LLaVA1.5-13B} & 34.4/59.6 & 8.8/89.6 & 4.8/88.0 & 12.0/84.8 & 1.6/96.4 & 12.8/82.0 & 9.6/87.6 & 31.2/62.8 & 14.4/81.3 \\
\textbf{OneVision-7B} & 32.0/36.4 & 21.6/54.8 & 2.8/94.4 & 24.8/56.4 & 2.4/86.4 & 30.0/38.0 & 11.6/71.2 & 42.4/31.2 & 20.9/58.6 \\
\textbf{Owl3-24-07} & 60.8/31.6 & 16.4/72.4 & 10.8/85.6 & 22.0/69.6 & 8.4/88.0 & 28.4/60.4 & 17.2/71.2 & 60.0/29.6 & 28.0/63.5 \\
\textbf{Qwen2VL-7B} & 26.4/58.0 & 12.4/82.8 & 0.8/95.6 & 13.2/80.4 & 4.0/93.6 & 16.0/78.8 & 11.6/83.6 & 38.0/54.0 & 15.3/78.3 \\
\textbf{Qwen2.5VL-7B} & 65.6/12.8 & 45.2/46.0 & 18.0/68.8 & 46.4/38.0 & 51.6/39.6 & 70.8/20.0 & 42.0/43.6 & 77.6/14.0 & 52.2/35.4 \\
\textbf{GLM-4V-9B} & 42.0/42.4 & 32.4/59.2 & 8.8/81.6 & 28.0/62.4 & 15.2/74.4 & 56.8/32.8 & 23.3/66.0 & 54.0/32.8 & 32.6/56.5 \\
\textbf{SPHINX-V2-1K} & 39.6/50.8 & 14.8/82.4 & 1.2/98.4 & 16.8/77.6 & 9.2/85.6 & 23.2/69.2 & 24.4/67.2 & 59.2/32.4 & 23.6/70.5 \\
\textbf{InternVL3-9B} & 45.2/35.2 & 21.2/68.0 & 20.8/62.4 & 27.2/54.4 & 23.2/50.4 & 38.0/40.4 & 19.6/63.2 & 76.8/14.8 & 34.0/48.6 \\		
\textbf{InternVL3-14B} & 72.8/8.8 & 30.8/48.4 & 25.2/60.0 & 33.2/52.0 & 37.2/47.2 & 58.0/21.2 & 24.8/52.8 & 84.4/9.6 & 45.8/37.5 \\
\textbf{CogVLM2-19B} & 44.0/39.6 & 29.2/56.0 & 8.8/75.6 & 19.2/54.8 & 8.0/73.2 & 31.6/43.2 & 25.2/60.8 & 59.2/28.4 & 28.2/54.0 \\
\textbf{InternVL3-38B} & 75.2/9.6 & 45.2/33.6 & 19.6/60.8 & 44.0/42.0 & 41.6/40.0 & 48.4/29.6 & 50.4/23.2 & 84.4/8.0 & 51.1/30.8 \\
\textbf{InternVL3-78B} & 92.4/3.2 & 46.0/28.8 & 66.4/18.4 & 41.6/37.2 & 69.6/13.2 & 76.4/8.8 & 74.4/12.8 & 89.6/4.0 & 69.5/15.8 \\
\textbf{Qwen2.5-VL-32B} & 85.60/10.40 & 49.20/39.20 & 49.60/42.80 & 52/37.60 & 52/42 & 70.80/20 & 57.20/35.20 & 86.80/10.40 & 62.90/29.70 \\
\textbf{Qwen2.5VL-72B} & 93.6/4.4 & 59.2/27.2 & 50.0/41.2 & 73.6/19.2 & 63.6/29.2 & 83.6/9.6 & 74.0/21.2 & 89.2/8.0 & 73.4/20.0 \\
\textbf{OneVision-72B} & 47.2/46.0 & 20.0/70.8 & 4.0/93.6 & 22.8/67.2 & 12.8/83.2 & 21.6/60.8 & 20.8/70.8 & 71.6/21.2 & 27.6/64.2 \\
\textbf{LLaVA1.6-7B} & 10.8/74.4 & 5.2/85.2 & 0.8/93.2 & 3.6/79.6 & 0.4/90.8 & 6.0/76.0 & 4.8/73.6 & 26.0/46.8 & 7.2/77.5 \\
\textbf{LLaVA1.6-13B} & 16.0/66.4 & 7.2/90.4 & 0.8/92.0 & 6.4/91.6 & 2.4/95.6 & 6.8/88.0 & 10.0/84.4 & 22.4/63.2 & 9.0/84.0 \\			
\textbf{LLaVA1.6-34B} & 34.8/42.4 & 12.0/81.6 & 6.8/85.6 & 16.8/76.0 & 11.2/83.2 & 25.2/60.8 & 14.0/76.0 & 60.0/31.6 & 22.6/67.2 \\

\midrule	
\textbf{GPT-4o-mini} & 94.4/3.2 & 35.6/47.6 & 60.4/28.4 & 22.0/58.9 & 19.4/59.2 & 34.8/36.4 & 71.2/20.4 & 78.4/12.8 & 52.0/33.4 \\
\bottomrule
\end{tabular}
}
\label{tab:single_full}
\end{table}

\begin{table}[ht]
\centering
\caption{Accuracy of question answering in the MC\textsuperscript{2} dataset when both textual and visual contexts are provided and the instruction explicitly directs the model to answer based on visual modality context. Values are reported as vision-based accuracy/text-based accuracy for each model.}
\resizebox{\linewidth}{!}{
\renewcommand\arraystretch{1.2}
\begin{tabular}{lccccccccc}
\toprule
\textbf{Model} & \textbf{Sport} & \textbf{Attribute} & \textbf{Sentiment} & \textbf{Positional} & \textbf{Counting} & \textbf{Color} & \textbf{Activity} & \textbf{Object} & \textbf{Avg} \\
\midrule
\textbf{OneVision-7B} &55.6/16.4 & 31.2/37.2 & 12.0/76.8 & 30.8/42.4 & 3.2/77.6 & 36.4/18.4 & 22.4/47.6 & 61.2/16.4 & 31.6/41.6  \\
\textbf{Qwen2VL-7B} &  60.8/26.8 & 24.0/69.2 & 20.0/69.6 & 20.4/74.0 & 10.8/80.0 & 32.0/52.0 & 27.2/61.6 & 63.2/28.8 & 32.3/57.8\\
\textbf{Qwen2.5VL-7B} & 77.6/14.4 & 43.2/46.8 & 18.4/72.8 & 43.2/40.4 & 35.6/55.6 & 58.8/24.4 & 53.6/35.6 & 81.2/11.6 & 51.4/37.7 \\
\textbf{CogVLM2-19B} & 73.2/13.2 & 47.6/32.4 & 35.6/28.4 & 26.8/45.2 & 14.0/40.0 & 61.6/17.6 & 56.0/28.0 & 76.0/15.2 & 48.9/27.5 \\
\textbf{InternLM-XC2.5-7B} & 84.0/9.6 & 46.4/42.8 & 74.0/18.4 & 36.0/52.4 & 22.8/66.0 & 63.6/20.8 & 74.0/18.4 & 76.4/15.6 & 59.7/30.5\\
\textbf{GLM-4V-9B} & 75.2/18.4 & 48.8/39.6 & 28.8/54.0 & 33.6/55.6 & 38.4/54.0 & 76.4/16.4 & 48.4/38.8 & 80.0/11.6 & 53.7/36.1 \\
\textbf{SPHINX-V2-1K} & 52.4/38.4 & 16.4/78.8 & 2.0/97.2 & 20.8/72.8 & 13.6/80.8 & 30.0/58.8 & 40.8/52.8 & 64.8/29.2 & 30.1/63.6 \\
\textbf{InternVL3-9B} & 96.0/2.0 & 67.2/18.8 & 82.8/13.2 & 54.8/26.4 & 55.6/21.2 & 84.4/7.6 & 82.8/6.4 & 91.6/4.0 & 76.9/12.4\\
\textbf{InternVL3-14B} & 98.4/0.8 & 86.0/4.4 & 87.6/7.6 & 71.6/12.8 & 78.0/6.8 & 97.2/0.8 & 90.8/3.2 & 96.4/1.6 & 88.2/4.8 \\
\textbf{LLaVA1.6-7B} & 33.2/54.0 & 6.8/80.8 & 6.0/82.4 & 6.4/79.6 & 2.8/90.4 & 10.8/70.0 & 13.6/69.2 & 48.4/40.8 & 16.0/70.9 \\
\textbf{LLaVA1.6-13B} &  41.6/40.4 & 10.4/85.2 & 4.0/62.8 & 8.4/83.6 & 5.6/92.8 & 14.4/70.8 & 24.8/58.0 & 45.2/41.2 & 19.3/66.9 \\
\textbf{LLaVA1.6-34B} &  84.8/12.0 & 48.0/36.4 & 62.8/24.0 & 34.0/52.0 & 38.8/44.4 & 76.4/14.4 & 62.0/18.4 & 80.4/12.4 & 60.9/26.8\\
\bottomrule
\end{tabular}
}

\label{tab:icc_img}
\end{table}

\begin{table}[ht]
\centering
\caption{Accuracy of question answering in the MC\textsuperscript{2} dataset when both textual and visual contexts are provided and the instruction explicitly directs the model to answer based on the textual modality. Values are reported as vision-based accuracy/text-based accuracy for each model.}
\resizebox{\linewidth}{!}{
\renewcommand\arraystretch{1.2}
\begin{tabular}{lccccccccc}
\toprule
\textbf{Model} & \textbf{Sport} & \textbf{Attribute} & \textbf{Sentiment} & \textbf{Positional} & \textbf{Counting} & \textbf{Color} & \textbf{Activity} & \textbf{Object} & \textbf{Avg} \\
\midrule
\textbf{OneVision-7B} & 45.6/16.4 & 22.4/37.2 & 5.6/76.8 & 27.2/42.4 & 1.6/77.6 & 28.8/18.4 & 16.8/47.6 & 56.0/16.4 & 25.5/41.6\\
\textbf{Qwen2VL-7B} &51.6/34.8 & 14.8/78.4 & 6.8/88.0 & 15.6/79.6 & 4.0/90.8 & 18.0/70.8 & 19.2/72.8 & 57.6/36.0 & 23.4/68.9 \\
\textbf{Qwen2.5VL-7B} & 77.6/14.4 & 43.2/46.8 & 18.4/72.8 & 43.2/40.4 & 35.6/55.6 & 58.8/24.4 & 53.6/35.6 & 81.2/11.6 & 51.4/37.7\\
\textbf{CogVLM2-19B} & 53.6/29.6 & 28.4/47.6 & 10.4/56.8 & 17.6/56.8 & 6.0/62.0 & 36.0/35.2 & 34.0/39.6 & 67.2/19.6 & 31.6/43.4 \\
\textbf{GLM-4V-9B} & 53.2/32.8 & 30.4/61.2 & 6.0/85.6 & 23.2/68.0 & 20.0/70.0 & 52.4/35.6 & 28.0/60.8 & 68.0/22.0 & 35.2/54.5 \\
\textbf{SPHINX-V2-1K} & 48.4/41.2 & 14.4/81.6 & 2.0/98.0 & 19.2/77.6 & 11.2/84.0 & 27.2/67.2 & 30.8/65.2 & 63.2/30.8 & 27.1/68.2 \\
\textbf{InternVL3-9B} &41.2/27.2 & 13.6/71.6 & 22.4/60.8 & 16.8/64.0 & 18.0/60.4 & 25.6/46.4 & 29.2/49.2 & 62.4/17.6 & 28.6/49.6 \\
\textbf{InternVL3-14B} & 28.4/44.8 & 14.0/68.4 & 3.6/82.4 & 21.2/54.8 & 28.0/43.2 & 24.8/50.0 & 17.2/58.8 & 55.2/19.6 & 24.0/52.8 \\
\bottomrule
\end{tabular}
}

\label{tab:icc_text}
\end{table}

\begin{table}[ht]
\centering
\caption{Accuracy of question answering in the MC\textsuperscript{2} dataset when only unimodal textual context is provided.}
\resizebox{\linewidth}{!}{
\renewcommand\arraystretch{1.2}
\begin{tabular}{lccccccccc}
\toprule
\textbf{Model} & \textbf{Sport} & \textbf{Attribute} & \textbf{Sentiment} & \textbf{Positional} & \textbf{Counting} & \textbf{Color} & \textbf{Activity} & \textbf{Object} & \textbf{Avg} \\
\midrule

\textbf{LLaMAVision} & 97.6 & 97.2 & 99.6 & 99.2 & 97.2 & 96.0 & 97.6 & 97.6 & 97.8 \\
\textbf{LLaVA1.5-7B} & 98.0 & 98.0 & 100.0 & 97.6 & 98.4 & 99.2 & 97.6 & 97.6 & 98.3 \\
\textbf{LLaVA1.5-13B} & 97.2 & 97.6 & 99.6 & 97.6 & 97.6 & 98.8 & 95.2 & 99.2 & 97.9 \\
 \textbf{OneVision-7B} & 98.0 & 95.2 & 100.0 & 98.4 & 98.0 & 98.8 & 98.0 & 100.0 & 98.3 \\
\textbf{Owl3} & 97.6 & 97.2 & 99.6 & 98.8 & 98.8 & 99.2 & 99.2 & 100.0 & 98.8 \\
\textbf{Qwen2VL-7B} & 98.8 & 96.4 & 99.6 & 99.6 & 98.8 & 100.0 & 98.8 & 100.0 & 99.0 \\
\textbf{Qwen2.5VL-7B} & 99.2 & 97.6 & 100.0 & 99.6 & 96.8 & 98.8 & 98.4 & 99.2 & 98.7 \\
\textbf{CogVLM2-19B} & 98.0 & 95.2 & 99.2 & 96.0 & 94.8 & 98.4 & 98.0 & 99.6 & 97.4 \\
\textbf{GLM-4V-9B} & 98.4 & 95.2 & 99.6 & 97.2 & 98.8 & 98.4 & 99.6 & 99.6 & 98.4 \\	
\textbf{SPHINX-V2-1K} & 98.4 & 97.6 & 99.2 & 98.8 & 98.0 & 99.2 & 98.4 & 99.6 & 98.7  \\

\textbf{InternVL3-9B} & 97.6 & 98.0 & 99.6 & 99.2 & 95.6 & 96.8 & 98.8 & 99.2 & 98.1\\
\textbf{InternVL3-14B} & 98.4 & 98.4 & 100.0 & 99.2 & 95.6 & 98.4 & 98.8 & 99.6 & 98.5 \\
\textbf{InternVL3-38B} & 97.6 & 96.8 & 100.0 & 98.8& 96.0 & 97.2 & 98.4 & 100.0 & 98.1 \\
\textbf{InternVL3-78B} & 97.2 & 97.6 & 100.0 & 98.0 & 96.4& 96.8 & 98.0 & 100.0 & 98.0 \\
\textbf{Qwen2.5VL-72B} & 99.6 & 98.4 & 96.8 & 100.0 & 97.2 & 100.0 & 99.6 & 99.2 & 98.9 \\
\textbf{OneVision-72B} & 100.0 & 97.6 & 97.6 & 99.6 & 96.4 & 100.0 & 100.0 & 98.8 & 98.7 \\
\midrule
\textbf{GPT-4o-mini} & 97.6 & 97.2 & 99.6 & 98.6 & 97.4 & 98.4 & 98.4 & 100.0 & 98.4\\
\bottomrule
\end{tabular}
}

\label{tab:base_text}
\end{table}

\begin{table}[ht]
\centering
\caption{Accuracy of question answering in the MC\textsuperscript{2} dataset when only unimodal visual context is provided.}
\resizebox{\linewidth}{!}{
\renewcommand\arraystretch{1.2}
\begin{tabular}{lccccccccc}
\toprule
\textbf{Model} & \textbf{Sport} & \textbf{Attribute} & \textbf{Sentiment} & \textbf{Positional} & \textbf{Counting} & \textbf{Color} & \textbf{Activity} & \textbf{Object} & \textbf{Avg} \\
\midrule
\textbf{LLaMAVision} & 100.0 & 98.8 & 92.8 & 98.4 & 96.4 & 99.2 & 98.8 & 97.2 & 97.7 \\
\textbf{LLaVA1.5-7B} & 99.6 & 98.0 & 96.4 & 100.0 & 97.6 & 99.6 & 98.8 & 98.4 & 98.5 \\
\textbf{LLaVA1.5-13B} & 99.6 & 95.2 & 94.4 & 97.6 & 95.2 & 98.4 & 96.4 & 98.4 & 96.9 \\
\textbf{OneVision-7B} & 100.0 & 97.2 & 97.2 & 98.4 & 84.4 & 99.6 & 97.2 & 98.8 & 96.6 \\
\textbf{Owl3} & 99.2 & 94.0 & 94.0 & 97.2 & 88.4 & 96.8 & 97.2 & 99.2 & 95.8 \\
\textbf{Qwen2VL-7B} & 99.6 & 98.8 & 95.6 & 98.4 & 96.4 & 100.0 & 99.6 & 98.4 & 98.3 \\
\textbf{Qwen2.5VL-7B} & 99.6 & 98.8 & 98.0 & 100.0 & 99.2 & 100.0 & 100.0 & 98.8 & 99.3 \\
\textbf{CogVLM2-19B} & 99.6 & 99.2 & 91.2 & 96.8 & 91.6 & 98.8 & 98.4 & 98.8 & 96.8 \\
\textbf{GLM-4V-9B} & 99.6 & 99.2 & 98.0 & 99.2 & 97.6 & 100.0 & 99.2 & 99.6 & 99.1 \\
\textbf{SPHINX-V2-1K} & 98.8 & 97.6 & 99.2 & 92.8 & 98.0 & 99.6 & 96.8 & 99.2 & 97.8 \\
\textbf{InternVL3-9B} & 98.8 & 95.6 & 95.6 & 96.8 & 90.0 & 100.0 & 98.0 & 98.0& 96.6 \\
\textbf{InternVL3-14B} & 99.2 & 96.4 & 96.4 & 98.4 & 92.4& 98.8 & 97.2 & 98.4 & 97.1 \\
\textbf{InternVL3-38B} & 100.0 & 98.0 & 97.2 & 100.0 & 94.4 & 99.6 & 99.2 & 98.8& 98.4 \\
\textbf{InternVL3-78B} & 99.2 & 99.6 & 96.8& 98.8 & 96.0 & 100.0 & 99.2 & 98.4 & 98.5 \\
\textbf{Qwen2.5VL-72B} & 97.2 & 97.2 & 100.0 & 99.2 & 97.2 & 98.4 & 98.4 & 99.6 & 98.4 \\
\textbf{OneVision-72B} & 100.0 & 97.6 & 97.6 & 99.6 & 96.4 & 100.0 & 100.0 & 98.8 & 98.7 \\
\midrule
\textbf{GPT-4o-mini} & 100.0 & 92.0 & 95.6 & 100.0 & 100.0 & 96.0 & 96.4 & 96.0 & 97.0 \\
\bottomrule
\end{tabular}
}

\label{tab:base_vision}
\end{table}